\documentclass[10pt,twocolumn,letterpaper]{article}

\usepackage[pagenumbers]{cvpr} %

\definecolor{cvprblue}{rgb}{0.21,0.49,0.74}
\usepackage{cuted}
\usepackage{capt-of}
\usepackage{tikz}
\usepackage[toc,page,header]{appendix}
\usepackage{minitoc}
\usepackage{pgfplots}
\usepackage[T1]{fontenc}
\usetikzlibrary{matrix}
\pgfplotsset{compat=1.18} %

\usepackage{url}

\newcommand{\gsc}{F}
\newcommand{\enc}{\mathcal{E}}
\newcommand{\dec}{\mathcal{D}}
\newcommand{\query}{\vec{q}}

\usepackage[hang,flushmargin]{footmisc} 

\usepackage[table]{xcolor}
\usepackage{multirow}
\usepackage{hhline}
\usepackage{booktabs}   
\usepackage{multirow}  
\usepackage{pifont}   
\usepackage{xcolor}     
\usepackage{array}     
\usepackage{booktabs}
\definecolor{refcolor}{gray}{0.9}
\definecolor{white}{gray}{1.0}
\newcommand{\cmark}{\textcolor{black}{\ding{51}}}
\newcommand{\xmark}{\textcolor{white!60!black}{\ding{55}}}
\usepackage{adjustbox}
\definecolor{lightpurblue}{RGB}{243, 249, 255}
\definecolor{midpurblue}{RGB}{205, 227, 255}
\definecolor{lightgreen}{RGB}{235, 255, 240}
\definecolor{midgreen}{RGB}{233, 253, 237}
\definecolor{lightred}{RGB}{254, 240, 236}
\newcommand{\best}[1]{\cellcolor{midpurblue}{#1}}
\newcommand{\secondbest}[1]{\cellcolor{lightpurblue}{#1}}
\newcommand{\secondbestdepth}[1]{\cellcolor{lightpurblue}{#1}}
\newcommand{\secondbesttrack}[1]{\cellcolor{lightpurblue}{#1}}

\newcommand{\para}[1]{\vspace{3pt} \noindent\textbf{#1}}
\usepackage{xspace}
\usepackage{multirow}
\usepackage{pifont}
\newcommand{\midsize}{\fontsize{9.5}{11.5}\selectfont}

\newcommand{\name}{\textsc{D4RT}\xspace}
\newcommand{\nameraw}{D4RT\xspace}
\newcommand{\task}{4D Reconstruction and Tracking\xspace}
\makeatletter
\newcommand*\bigcdot{\mathpalette\bigcdot@{.8}}
\newcommand*\bigcdot@[2]{\mathbin{\vcenter{\hbox{\scalebox{#2}{$\m@th#1\bullet$}}}}}
\makeatother
\renewcommand{\vec}[1]{\mathbf{#1}}

\usepackage{algorithm}
\usepackage{algpseudocode}

\renewcommand{\Comment}[1]{\hfill\textcolor{gray}{\(\triangleright\)\ \scriptsize{#1}}}

\usetikzlibrary{tikzmark,calc}

\usepackage{setspace}

\newcommand\blfootnote[1]{%
  \begingroup
  \renewcommand\thefootnote{}\footnote{#1}%
  \addtocounter{footnote}{-1}%
  \endgroup
}

\usepackage[pagebackref,breaklinks,colorlinks,allcolors=cvprblue]{hyperref}

\newcommand{\authorspace}{\quad}
\newcommand{\gdm}{\textsuperscript{$\star$}}

\newcommand{\ucl}{\textsuperscript{$\diamond$}}
\newcommand{\oxf}{\textsuperscript{$\circ$}}

\title{Efficiently Reconstructing Dynamic Scenes One {\raisebox{-0.17em}{\includegraphics[height=1em]{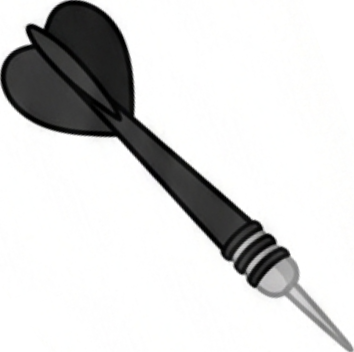}}} \name at a Time}

\author{
    Chuhan Zhang\gdm
    \authorspace 
    Guillaume Le Moing\gdm
    \authorspace 
    Skanda Koppula\gdm\ucl
    \authorspace 
    Ignacio Rocco\gdm
    \\
    Liliane Momeni\gdm
    \authorspace
    Junyu Xie\oxf\textsuperscript{1}
    \authorspace
    Shuyang Sun\gdm
    \authorspace 
    Rahul Sukthankar\gdm
    \authorspace
    Joëlle K. Barral\gdm
    \\ 
    Raia Hadsell\gdm
    \authorspace
    Zoubin Ghahramani\gdm
    \authorspace
    Andrew Zisserman\gdm\oxf
    \authorspace 
    Junlin Zhang\gdm
    \\
    Mehdi S. M. Sajjadi\gdm\textsuperscript{2}
    \vspace{4mm}
    \\
    {\midsize\gdm Google DeepMind} \quad 
    {\midsize\ucl University College London} \quad
    {\midsize\oxf University of Oxford}\vspace{-3mm}
}

\begin{document}

\maketitle

{
\begin{abstract}
    Understanding and reconstructing the complex geometry and motion of dynamic scenes from video remains a formidable challenge in computer vision.
    This paper introduces \name, a simple yet powerful feedforward model designed to efficiently solve this task.
    \name utilizes a unified transformer architecture to jointly infer depth, spatio-temporal correspondence, and full camera parameters from a single video.
    Its core innovation is a novel querying mechanism that sidesteps the heavy computation of dense, per-frame decoding and the complexity of managing multiple, task-specific decoders.
    Our decoding interface allows the model to independently and flexibly probe the 3D position of any point in space and time.
    The result is a lightweight and highly scalable method that enables remarkably efficient training and inference.
    We demonstrate that our approach sets a new state of the art, outperforming previous methods across a wide spectrum of 4D reconstruction tasks.
    We refer to the project webpage for animated results.\textsuperscript{3}
\end{abstract}

\section{Introduction}
\label{sec:intro}

Traditional 3D reconstruction asks:
\emph{`What is the geometry of everything, everywhere, \underline{all at once}?'}
We argue this exhaustive, rigid approach is fundamentally ill-equipped for a dynamic world.
Despite the clear need for unified 4D understanding, leading approaches often tackle the problem by dividing it into discrete, task-specific components.
\blfootnote{
\textsuperscript{1}Work was done during an internship at Google DeepMind.\\
\textsuperscript{2}Correspondence: \url{d4rt@msajjadi.com}\\
\textsuperscript{3}Project website: \url{https://d4rt-paper.github.io/}
}

For instance, MegaSaM~\cite{li2025megasam} relies on a complex mosaic of off-the-shelf models to separately estimate mono-depth, metric depth, and motion segmentation.
Fusing these disparate signals requires computationally expensive test-time optimization to enforce geometric consistency.
Recent feedforward approaches such as VGGT~\cite{wang2025vggt} employ separate, specialized decoders for distinct modalities.
Crucially, neither of these methods is capable of establishing correspondences for \emph{dynamic} portions of the scene.
While SpatialTrackerV2~\cite{xiao2025spatialtrackerv2} incorporates dynamics, it still lacks a unified, single-stage formulation, instead relying on costly iterative refinement.
\begin{figure}
\centering
\includegraphics[width=1.0\linewidth]{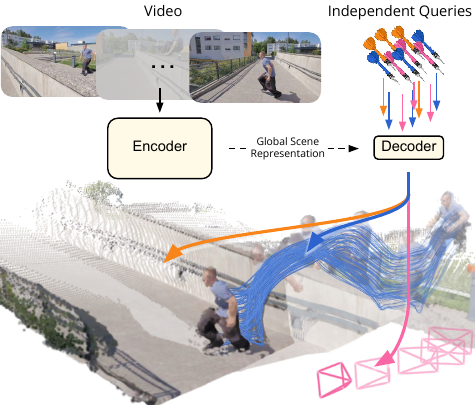}
\captionof{figure}{
\textbf{\name} is a unified, efficient, feedforward method for \emph{\textbf{D}ynamic \textbf{4}D \textbf{R}econstruction and \textbf{T}racking}, unlocking a variety of outputs including point cloud ({\raisebox{-0.25em}{\includegraphics[height=1.2em]{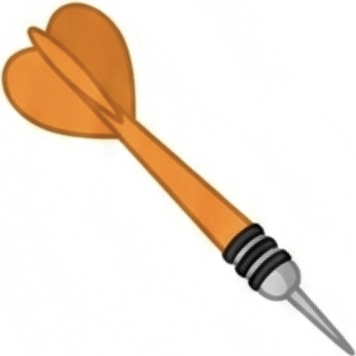}}}), point tracks ({\raisebox{-0.25em}{\includegraphics[height=1.2em]{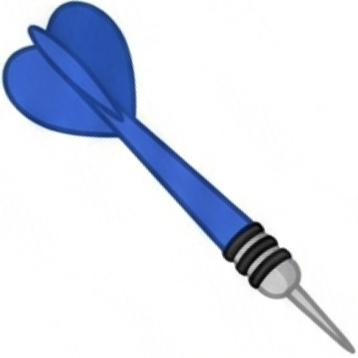}}}), camera parameters ({\raisebox{-0.25em}{\includegraphics[height=1.2em]{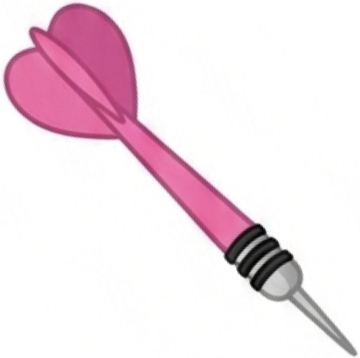}}}) through a single interface.
}
\label{fig:teaser}
\end{figure}

\begin{figure*}
\centering
\includegraphics[width=1.0\linewidth]{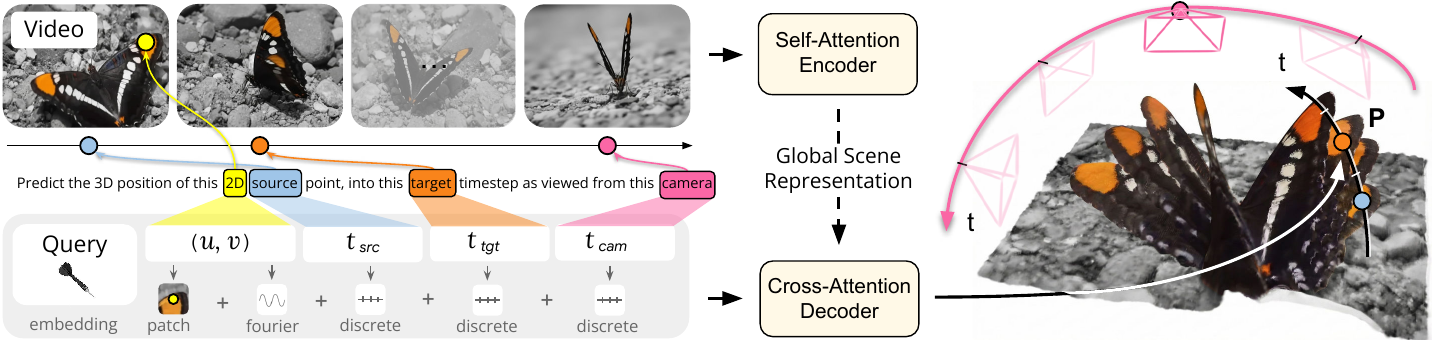}
\caption{
\textbf{\name model overview} --
A global self-attention encoder first transforms the input video into the latent \emph{Global Scene Representation} $\gsc$, which is passed to a lightweight decoder.
The decoder can be independently queried for the \emph{3D position} $\vec{P}$ of any given 2D point ($u$, $v$) from the source timestep $t_\text{src}$ at target timestep $t_\text{tgt}$ in camera coordinate $t_\text{cam}$, unlocking full decoding at any point in space and time.
The query also contains an embedding of the local video patch centered around ($u$, $v$), providing additional spatial context.
}
\label{fig:method}
\end{figure*}

We propose shifting the paradigm from fragmented, frame-level decoding to efficient, on-demand querying.
We introduce \name, a feedforward method leveraging a flexible and scalable architecture to achieve full 4D reconstruction.
As shown in \cref{fig:teaser}, our model first encodes the input video, generating a latent scene representation which is then used to independently decode any number of spatio-temporal point queries.
This simple, novel design unifies all 4D reconstruction tasks into a single interface and unlocks efficient training and inference.

\noindent Our key contributions are as follows:
\begin{itemize}
    \item We propose \name, a novel method for efficient feedforward querying of point-level 4D scene information captured in a video.
    \item We demonstrate how our unified approach unlocks 4D correspondence, point clouds, depth maps, and camera parameters for both static and dynamic scenes through a single interface.
    \item In an extensive set of experiments, we show that \name sets a new state of the art in
    \emph{Dynamic 4D Reconstruction and Tracking} while outperforming existing approaches in both speed and accuracy.
    \item Finally, we demonstrate how our flexible decoder unlocks an efficient algorithm to track all pixels in a video, enabling dense, holistic scene reconstruction.
\end{itemize}

\section{Method}
\label{sec:method}

\name is based on a simple encoder-decoder architecture inspired by the Scene Representation Transformer~\cite{srt, rust}.
As shown in \cref{fig:method}, the video is first processed by a powerful encoder producing the \emph{Global Scene Representation} $\gsc$.
The role of the encoder is to capture information about the full environment, identifying dense correspondence across all video frames, as well as understanding the flow of time and its effect on the scene.
In a second stage, a lightweight decoder queries $\gsc$ through a simple low-level interface.
Specifically, given a 2D point in a \emph{source frame}, the decoder predicts the \emph{3D position} of the point at a given \emph{target timestep} (defining the temporal state) and expressed relative to a given \emph{camera} reference (defined by the frame timestep where the camera viewpoint corresponds to this reference).

We draw attention to three desirable properties of this formulation:
first, the indices need not coincide, allowing a full disentanglement of space and time;
second, each query is decoded independently, allowing for both efficient training and inference as well as flexible decoding (both sparse and dense); and
third, this interface unlocks a suite of downstream applications in a unified, consistent manner (\cref{tab:query_mechanism}).

\subsection{\nameraw Framework}
\label{sec:method:framework}

Given a video $V \in \mathbb{R}^{T\times H\times W\times 3}$, the encoder $\enc$ extracts the latent \emph{Global Scene Representation}
\begin{equation*}
\gsc = \enc(V) \in \mathbb{R}^{N\times C}.
\end{equation*}

Once $\gsc$ is calculated, it remains fixed throughout the second stage, where the decoder $\dec$ cross-attends from any number of queries into $\gsc$.
We define a query $\query = (u, v, t_\text{src}, t_\text{tgt}, t_\text{cam})$, where $(u, v, t_{src})$ correspond to \emph{source} parameters and $(t_\text{tgt}, t_\text{cam})$ correspond to \emph{target} parameters.
Here, $(u, v) \in [0, 1]^2$ represent the normalized 2D coordinates of a point of interest in the source frame $t_{src}$, while $(t_\text{tgt}, t_\text{cam}) \in [1,\ldots, T]$ denote the temporal indices of the target timestep and the reference camera coordinate system (illustrated in \cref{fig:method}).
Each query $\query$ is processed \emph{fully independently} with the video features $\gsc$ to produce its corresponding 3D point position $\vec{P}$:
\begin{equation*}
\vec{P} = \dec(\query, \gsc) \in \mathbb{R}^3.
\end{equation*}

\para{From query to 4D reconstruction.}
Through a simple variation of queries, our framework allows us to address a broad range of 4D tasks as shown in \cref{tab:query_mechanism}.
Choosing any fixed point $(u, v)$ from a source frame $t_\text{src}$ in the video while varying $t_\text{tgt} = t_\text{cam} = \{1\ldots T\}$ produces its \emph{point track}, the 3D trajectory of the corresponding point throughout the video.
For full \emph{point cloud} reconstruction, the 3D position of all pixels in the video can be directly predicted in a shared reference frame $t_\text{cam}$ by the model.
This alleviates the need for coordinate transformations to map pixels from different video frames
into a unified coordinate system using explicit, potentially noisy camera estimates.
\emph{Depth maps} can be recovered by simply querying any pixel in the video with $t_\text{src}=t_\text{tgt}=t_\text{cam}$ and only keeping the Z-dimension of the output $\vec{P}$.

We next detail how \emph{camera extrinsics} and \emph{intrinsics} predictions are obtained.
To derive the relative camera pose between any pair of video frames $i,j\in[1\ldots T]$, we create queries for an ensemble of source points $\{(u_k, v_k)\}_k$ sampled on a $(h,w)$ grid in both reference frames:
\begin{equation*}
\query_{i,k} = (u_k, v_k, i, i, i),\quad
\query_{j,k} = (u_k, v_k, i, i, j).
\end{equation*}
The resulting sets $\{\dec(\query_{i,k}, \gsc)\}_k$ and $\{\dec(\query_{j,k}, \gsc)\}_k$ describe the same 3D points in different reference frames.
We therefore only need to find the rigid transformation between them, which can be efficiently derived through Umeyama's algorithm~\cite{umeyama1991least} that solves a $3{\times}3$ SVD decomposition.

To recover intrinsics for video frame $i\in[1\ldots T]$, we construct a set of queries for different source points, again sampled on a $(h,w)$ grid.
We decode all corresponding 3D positions $\vec{P}=(p_x, p_y, p_z)$.
Assuming a pinhole camera model with a principal point at $(0.5, 0.5)$, we get focal length parameters as follows:
\begin{equation*}
    f_x {=} p_z(u - 0.5)/p_x,\quad
    f_y {=} p_z(v - 0.5)/p_y.
\end{equation*}
We take the median over the $k$ estimates for robustness.
Camera models with distortion (\eg, fisheye) can also be seamlessly incorporated by adding a non-linear refinement step on top of the initial estimation~\cite{zhang2002flexible}.

\newcommand{\klghfrskfh}{\hhline{>{\arrayrulecolor{black}}------}}
\newcommand{\ihatelatex}{\hhline{>{\arrayrulecolor{white}}-----}}

\begin{table}[tb!]
\hspace*{-0.033\linewidth}
\resizebox{1.045\linewidth}{!}{
\begin{small}
    \centering
    \setlength{\tabcolsep}{6pt}
\begin{tabular}{@{}l 
  c !{\color{white}\vrule width 2pt} %
  c !{\color{white}\vrule width 2pt} %
  >{\centering\arraybackslash}c 
  !{\color{white}\vrule width 2pt} %
  >{\centering\arraybackslash}c 
  !{\color{white}\vrule width 2pt} %
  >{\centering\arraybackslash}c 
}

\klghfrskfh\klghfrskfh\\[-1em]
\multirow{2}{*}{\textbf{~~~~~~~~~~~~Task}} & \multicolumn{5}{c}{\textbf{Query}} \\ 

\hhline{~>{\arrayrulecolor{black}}-----} 
\hhline{~>{\arrayrulecolor{white}}-----}

& $u$ & $v$ & $t_{src}$ & $t_{tgt}$ & $t_{cam}$ \\
\\[-1em] 

{\raisebox{-0.25em}{\includegraphics[height=1.2em]{imgs/darts/blue_down.png}}} ~ Point Track 
& \cellcolor{refcolor}{\scriptsize Fixed} & \cellcolor{refcolor}{\scriptsize Fixed} 
& \cellcolor{refcolor} {\scriptsize Fixed} 
& \multicolumn{2}{c}{\cellcolor{refcolor} {\scriptsize$1$\ldots$T$}} \\ 
\\[-1em] 

{\raisebox{-0.25em}{\includegraphics[height=1.2em]{imgs/darts/orange_down.png}}} ~ Point Cloud 
& \cellcolor{refcolor}{\scriptsize$1$\dots$W$} & \cellcolor{refcolor}{\scriptsize$1$\dots$H$} 
& \multicolumn{2}{c|@{\tiny ~}}{\cellcolor{refcolor} {\scriptsize$1$\ldots$T$}} 
& \cellcolor{refcolor} {\scriptsize Fixed} \\ 
\\[-1em] 

{\raisebox{-0.25em}{\includegraphics[height=1.2em]{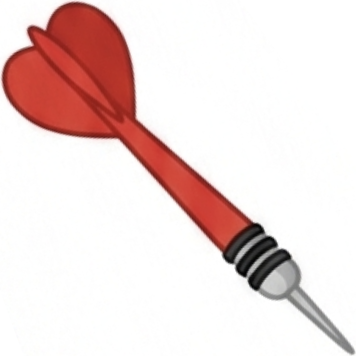}}} ~ Depth Map 
& \cellcolor{refcolor}{\scriptsize$1$\dots$W$} & \cellcolor{refcolor}{\scriptsize$1$\dots$H$} 
& \multicolumn{3}{c}{\cellcolor{refcolor} {\scriptsize$1$\ldots$T$}} \\ 
\\[-1em] 

{\raisebox{-0.25em}{\includegraphics[height=1.2em]{imgs/darts/pink_down.png}}} ~ Extrinsics 
& \cellcolor{refcolor}{\scriptsize$1...h$} & \cellcolor{refcolor}{\scriptsize$1...w$} 
& \multicolumn{2}{c|@{\tiny ~}}{\cellcolor{refcolor} {\scriptsize Fixed}} 
& \cellcolor{refcolor} {\scriptsize$1$\ldots$T$} \\ 
\\[-1em] 

{\raisebox{-0.25em}{\includegraphics[height=1.2em]{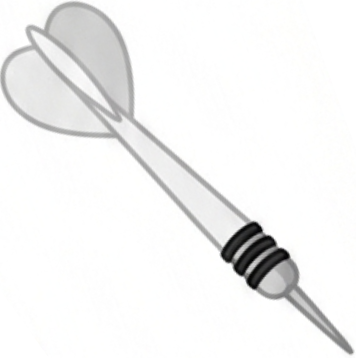}}} ~ Intrinsics 
& \cellcolor{refcolor}{\scriptsize$1...h$} & \cellcolor{refcolor}{\scriptsize$1...w$} 
& \multicolumn{3}{c}{\cellcolor{refcolor} {\scriptsize$1$\ldots$T$}} \\ 
\\[-1em] 
\klghfrskfh

\end{tabular}

\end{small}
}

\caption{
    \textbf{Unified decoding} --
    A diverse set of geometry-related tasks can be inferred by querying the Cartesian product of the respective entries.
    Note that for intrinsics and extrinsics, we only query a coarse $(h,w)$ grid for faster inference.
}

\label{tab:query_mechanism}
\end{table}

\subsection{Model Architecture}
\label{sec:method:architecture}

Our encoder $\enc$ is based on the Vision Transformer~\cite{dosovitskiy2021} with interleaved \emph{local frame-wise}, and \emph{global} self-attention layers~\cite{wang2025vggt}.
For simplicity, and to support arbitrary aspect ratios, we resize the input video to a fixed square resolution before tokenizing it.
To incorporate the original aspect ratio, we embed it into a separate token and pass it to the transformer along with the main video tokens.

\para{Pointwise decoder.}
The decoder $\dec$ is a small cross-attention transformer.
A query token is first constructed by adding the Fourier feature embedding~\cite{AttentionIsAllYouNeed} of the 2D coordinates $(u, v)$ to the learned discrete timestep embeddings for $t_\text{src}$, $t_\text{tgt}$, and $t_\text{cam}$.
We empirically observe that augmenting the query with an embedding of the local $9{\times}9$ pixel RGB patch centered at $(u, v)$ dramatically improves performance, see \cref{sec:exp:ablations}.

Each query is decoded independently through cross-attention into the \emph{Global Scene Representation} $\gsc$, ensuring that queries do not interact.
The resulting output feature is then mapped to a 3D point position $\vec{P}$ via a simple learned projection.
Decoding queries independently is a deliberate design decision with major advantages.
It allows efficient training, as only a small number of queries need to be decoded to provide a supervision signal to the model.
Equally importantly, at inference time, the queries can be chosen freely across all video frames as described in \cref{sec:method:training}, and need not be correlated to each other to avoid out-of-distribution effects -- indeed, we have empirically observed major performance drops when enabling self-attention between queries in early experiments.
Finally, this design allows highly efficient inference thanks to its trivial parallelism.
As shown in \cref{fig:speed} and \cref{tab:fps}, we obtain the best trade-off between performance and latency across multiple tasks.

\newcommand{\grid}{G}
\begin{algorithm}[t]
\small
\begin{algorithmic}[1]
\Require Input video $V$, model encoder $\enc$ and decoder $\dec$.
\State $\gsc \gets \enc(V)$ \Comment{Compute \emph{Global Scene Representation}}
\State $\grid \gets \{ \textbf{false} \}^{T \times H \times W}$ \Comment{Initialize \emph{Occupancy Grid}}
\State $\mathcal{T} \gets \emptyset$ \Comment{Initialize \emph{Set of Dense Tracks}}
\While{\textbf{any}($\grid = \textbf{false}$)}
    \State Sample a batch $B$ of unvisited source points from $\grid$
    \For{each $(u, v, t_{\text{src}}) \in B$} \Comment{Process batch in parallel}
        \State $Q \gets \{u, v, t_\text{src}, {t_\text{tgt}{=}t_\text{cam}}{=}k\}_{k=1}^T$ \Comment{Get \emph{Track Queries}}
        \State $P \gets  \{\dec(\query_k, \gsc)\}_{k=1}^T$  \Comment{Run the decoder}
        \State $\grid \gets \text{Visible}(P) $ \Comment{Set visible track pixels as visited}
        \State $\mathcal{T} \gets \mathcal{T} \cup P$ \Comment{Add new track to the output}
    \EndFor
\EndWhile
\State \textbf{return} $\mathcal{T}$
\end{algorithmic}
\caption{Efficient Dense Tracking of All Pixels.}
\label{alg:track_all_pixels}
\end{algorithm}

\subsection{Training and Inference}
\label{sec:method:training}

The model is implemented in Kauldron~\cite{kauldron2025github} and trained end-to-end by minimizing the weighted sum of losses computed over a batch of $N$ sampled queries.
The primary supervision signal is derived from an $L_1$ loss applied to the normalized 3D point position $\vec{P}$.
Specifically, both the target and the estimated point sets are normalized by their respective mean depths~\cite{wang2024dust3r} and then passed through the transform $\text{sign}(x) \cdot \log(1 {+} |x|)$ to dampen the influence of far-away points on the loss.
We also supervise a set of auxiliary predictions from additional linear projections on the decoder output:
An $L_1$ loss on \emph{2D coordinates} of the point positions in image space;
cosine similarity for \emph{3D surface normals}~\cite{wang2025pi};
binary cross-entropy for \emph{target point visibility}; and $L_1$ on the vector of \emph{point motion}.
All loss terms are applied only where ground truth supervision is available.
We finally incorporate a confidence penalty $\text{--}\log(c)$ where $c$ additionally weights the 3D point error~\cite{kendall2017uncertainties}.

\para{Efficient dense dynamic correspondence.}
A key capability of our query-based model is that it can efficiently compute dense correspondences for \emph{all} pixels in a video, both static and dynamic.
As shown in \cref{fig:point_cloud_vis}, this capability is crucial for building a complete, holistic scene reconstruction, which in turn is key to eliminating the occlusion-induced discontinuities and sparse artifacts common in prior works.
However, a naive approach to reconstruct tracks for all pixels across the video would involve $O(T^2 H W)$ queries, the majority of which are not required.

We introduce \cref{alg:track_all_pixels} which exploits spatio-temporal redundancy using an occupancy grid $\grid \in \{0, 1\}^{T \times H \times W}$ to speed this procedure up significantly.
The algorithm only initiates new tracks from unvisited pixels.
Each \emph{full-video track} marks all spatio-temporal pixels it \emph{visibly} passes through as visited.
Empirically, we find that this yields an adaptive speedup between 5--15$\times$ depending on the motion complexity in the video.
This dense, flexible strategy is feasible for our method precisely because our decoder is \emph{both} sparse, and lightweight.
It is notable that prior works are ill-suited for this task for various reasons.
Most methods fail to provide any correspondences for dynamic portions of the scenes~\cite{wang2025vggt, wang2025pi, li2025megasam}, dense frame-level decoding remains locked in the costly naive approach~\cite{feng2025st4rtrack}, and models with heavy sparse decoders face a large per-query cost~\cite{karaev2025cotracker3, xiao2025spatialtrackerv2}.

\begin{table}[t]
\centering
\small
\tabcolsep=0.2em
\renewcommand{\arraystretch}{1.1}

\resizebox{\linewidth}{!}{
\begin{tabular}{@{}l cc cccc} 
\toprule
\multirow{2}[6]{*}{\textbf{Model}} & \multicolumn{2}{c}{\textbf{Task}} & \multicolumn{2}{c}{\textbf{Functionality}}   & \multicolumn{2}{c}{\textbf{Architecture}}\\ 
\cmidrule(lr){2-3} \cmidrule(lr){4-5}  \cmidrule(lr){6-7}
 & {\scriptsize 3D Recon-} & {\scriptsize Dynamic Cor-} & {\scriptsize Flexible} & {\scriptsize Sparse} & {\scriptsize Global} & {\scriptsize Single} \\ [-0.6em]
 & {\scriptsize struction} & {\scriptsize respondence} & {\scriptsize Ref. Frames} & {\scriptsize Decoding} & {\scriptsize Context} & {\scriptsize Decoder} \\ 
\midrule
MegaSaM~\cite{li2025megasam} & \cmark & \xmark & \xmark & \xmark & \xmark & \xmark \\ 
DUSt3R\cite{wang2024dust3r} & \cmark & \xmark & \xmark & \xmark & \xmark & \xmark \\ 
VGGT~\cite{wang2025vggt} & \cmark & \xmark & \xmark & \xmark & \cmark & \xmark \\ 
$\pi^3$~\cite{wang2025pi}& \cmark & \xmark & \cmark & \xmark & \cmark & \xmark \\ 
St4RTrack~\cite{feng2025st4rtrack} & \cmark &  \cmark & \xmark & \xmark & \xmark &  \xmark \\ 
STv2~\cite{xiao2025spatialtrackerv2} & \cmark & \cmark & \xmark & \cmark & \cmark & \xmark \\ 
\midrule
\textbf{\name (Ours)} & \cmark & \cmark & \cmark & \cmark & \cmark & \cmark \\ 
\bottomrule
\end{tabular}
}
\caption{\textbf{Model capabilities} --
We highlight both the tasks our model executes but also its comprehensive functionality and simple model architecture.
}
\label{tab:capabilities}
\end{table}

\begin{figure}[t]

\includegraphics[trim={2.1cm 22cm 11.2cm 2.5cm},clip,width=1.0\linewidth]{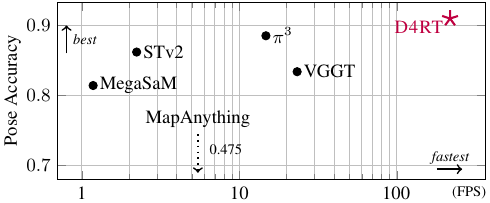}

\caption{
    \textbf{Pose accuracy \vs speed} -- 
    We compare pose accuracy \vs throughput against recent state-of-the-art methods.
    Pose accuracy is \emph{1 -- error}, averaged over ATE/RTE/RPE on Sintel and ScanNet.
    Throughput is measured in FPS on an A100 GPU.
    \name achieves 200+ FPS pose estimation, ~9$\times$ faster than VGGT, and 100$\times$ faster than MegaSaM, while delivering superior accuracy.
}
\label{fig:speed}
\end{figure}

\section{Related Work}
\label{sec:related}

Classical approaches to 3D reconstruction are fundamentally anchored in multi-view geometry, specifically Structure-from-Motion (SfM)~\cite{hartley2003multiple, oliensis2000critique} and Multi-View Stereo (MVS)~\cite{furukawa2015stereo,schonberger2016pixelwise}.
The standard pipeline exemplified by COLMAP~\cite{schonberger2016structure} incrementally estimates and optimizes sparse geometry and camera poses.
The advantage of these methods is their explicit enforcement of geometric consistency and mathematically interpretable reconstructions.
However, they are computationally intensive and brittle.

\subsection{Feedforward 3D Reconstruction}
\label{sec:related:recon}
The field of 3D reconstruction has recently shifted towards end-to-end, feedforward models that directly infer geometry from images.
The seminal work DUSt3R~\cite{wang2024dust3r} 
demonstrated that Transformer-based networks~\cite{AttentionIsAllYouNeed} can perform 3D reconstruction from unposed and uncalibrated image pairs in an end-to-end approach.
VGGT~\cite{wang2025vggt} later scaled this approach beyond pairs by using a Vision Transformer~\cite{dosovitskiy2021} with global attention.
Building on this feedforward paradigm, several works have extended the methodology to dynamic videos~\cite{zhang2024monst3r, wang2025cut3r, wang2025c4d, chen2025easi3r} and more efficient inference~\cite{wang2025pi, yang2025fast3r}.
However, these models share significant limitations.
For instance, many do not support changing camera intrinsics within a video~\cite{wang2025pi, wang2025vggt, li2025megasam}, and several are restricted to using only the first frame as the camera reference~\cite{wang2024dust3r,li2025megasam, wang2025cut3r}.
More importantly, architectures inspired by VGGT~\cite{wang2025vggt} either use separate decoder heads for each task~\cite{zhang2024monst3r, wang2025cut3r, wang2025pi} (\eg, depth, pose, point cloud), or they incorporate separate models for sub-tasks of 4D reconstruction~\cite{wang2025c4d,lu2025align3r, sucar2025dynamic}, making the full pipeline cumbersome and computationally expensive to run.
Finally, these methods are not directly capable of providing correspondences for dynamic regions of the scene.

\begin{figure*}[t]
\centering
\small
\tabcolsep=0em
\noindent\begin{tabular}{
  >{\centering\arraybackslash}p{0.095\linewidth}
  >{\centering\arraybackslash}p{0.2275\linewidth}
  >{\centering\arraybackslash}p{0.2272\linewidth}
  >{\centering\arraybackslash}p{0.2255\linewidth}
  >{\centering\arraybackslash}p{0.2255\linewidth}
}
{Input video} & ~~~~~~MegaSaM~\cite{li2025megasam} & ~~~$\pi^3$~\cite{wang2025pi} & ~~SpatialTrackerV2~\cite{xiao2025spatialtrackerv2} & \textbf{D4RT (Ours)} \\[5pt]
\end{tabular}
\includegraphics[width=0.98\linewidth, trim=0.2cm 0 0 1cm, clip]{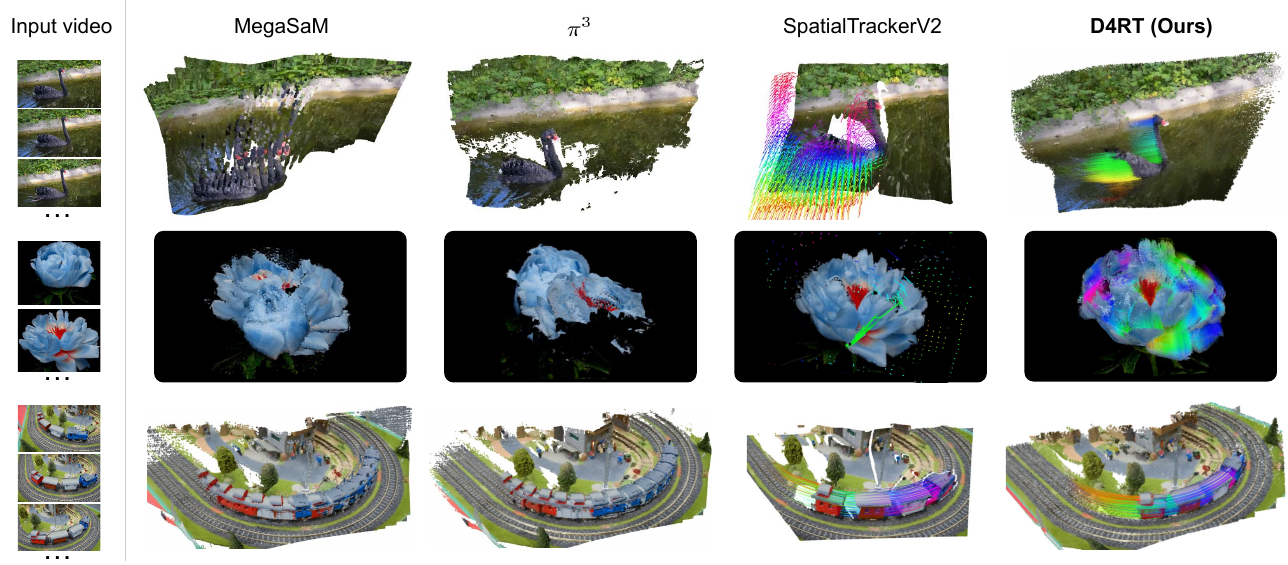}
\caption{
\textbf{Reconstruction results across methods} --
Pure reconstruction methods (MegaSaM and $\pi^3$) are only able to accumulate point clouds of all pixels; exhibiting clear failure cases in dynamic scenes.
For example, the swan is repeated in MegaSaM's reconstruction,
and $\pi^3$ is failing entirely to reconstruct the flower.
SpatialTrackerV2, a state-of-the-art tracking method, successfully captures dynamics, however its design only allows tracking points from \emph{one} frame, leaving gaps in the reconstruction (behind the swan and train).
\name is the only method that successfully reconstructs a full 4D representation of the scene including \emph{all} pixels of the video.
}
\label{fig:point_cloud_vis}
\end{figure*}

\subsection{From 2D to 3D Point Tracking}
\label{sec:related:tracking}

The task of point tracking aims to establish long-term 2D correspondences through challenges like occlusions and non-rigid motion, evolving from early methods such as Particle Video~\cite{sand2008particle} to newer deep-learning approaches~\cite{harley2022particle, doersch2022tap, doersch2023tapir, zholus2025tapnext}.
This field has further progressed from tracking a sparse set of points to the dense tracking of every pixel in a video~\cite{le2024dot, ngo2024delta, harley2025alltracker}.
A parallel line of work has explored lifting these 2D tracks into 3D, which is achieved by projecting tracks from image space to camera coordinates using ground truth or predicted depth and camera intrinsics~\cite{koppula2024tapvid}.

\begin{figure*}[h]
\centering
\includegraphics[width=\linewidth]{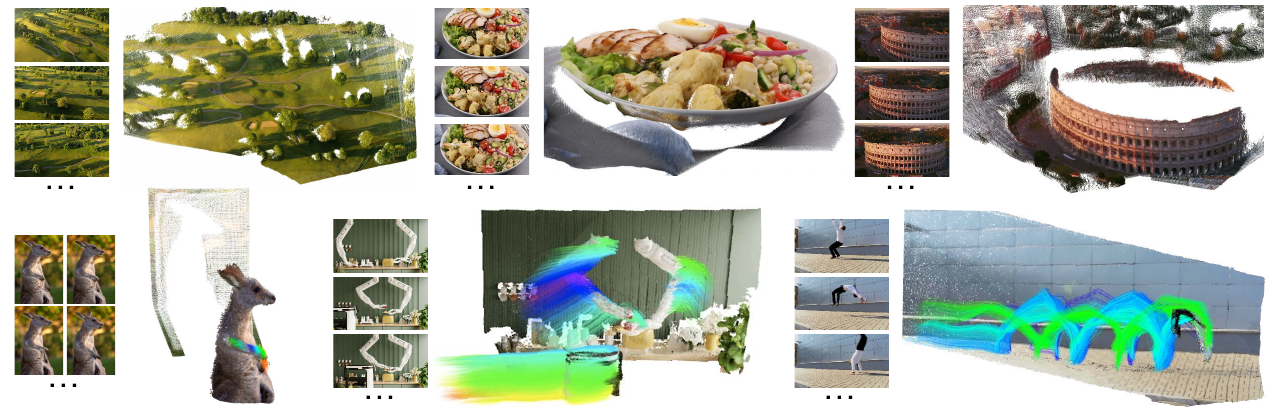}
\caption{
\textbf{Visualizations on in-the-wild videos} -- \name demonstrates accurate reconstructions on static (top row) and dynamic scenes (bottom row).
In the presence of motion, \name additionally produces robust 3D point trajectories.
}
\label{fig:ours_vis}
\end{figure*}

Most recently, models like SpatialTracker~\cite{xiao2024spatialtracker, xiao2025spatialtrackerv2}, L4P~\cite{badki2026l4p}, DPM~\cite{sucar2025dynamic} and St4RTracker~\cite{feng2025st4rtrack} have moved to directly predicting 3D tracks without priors. Yet, limitations remain: St4RTracker~\cite{feng2025st4rtrack} and DPM~\cite{sucar2025dynamic} follow the pairwise paradigm of DUSt3R~\cite{wang2024dust3r}, preventing holistic video processing. SpatialTrackerV2~\cite{xiao2025spatialtrackerv2} is a multi-stage approach and exhibits slow inference speeds as it relies on iterative refinements. Meanwhile, L4P~\cite{badki2026l4p} requires different heads for sparse and dense outputs of the same nature (\eg, flow vs. 2D tracking or depth vs. 3D tracking). In contrast, our approach unifies these tasks within a single architecture.
\cref{tab:capabilities} summarizes capabilities across a recent set of state-of-the-art models.

\section{Experiments}
\label{sec:exp}

We benchmark \name against recent state-of-the-art methods.
After inspecting qualitative differences in capability in \cref{sec:exp:qualitative}, we proceed with evaluating \task performance in \cref{sec:exp:4d}, focusing the comparison on the best methods from \cref{sec:related:tracking}.
We then continue with pure reconstruction tasks in \cref{sec:exp:3d}, comparing to the recent SOTA models mentioned in \cref{sec:related:recon}.
We conclude with a set of key ablations in \cref{sec:exp:ablations}.

\para{Training setup.}
For the encoder, we use the ViT-g model variant with 40 layers on a spatio-temporal patch size of $2{\times}16{\times}16$.
This encoder and our 8-layer cross-attention decoder contain 1\,B and 144\,M parameters, respectively.
Our training mixture contains public and internal datasets including
BlendedMVS~\cite{yao2020blendedmvs},
Co3Dv2~\cite{reizenstein2021common},
Dynamic Replica~\cite{karaev2023dynamicstereo},
Kubric~\cite{movi-e},
MVS-Synth~\cite{DeepMVS},
PointOdyssey~\cite{PointOdyssey}, ScanNet++~\cite{yeshwanthliu2023scannetpp},
ScanNet~\cite{dai2017scannet},
Tartanair~\cite{tartanair2020iros},
VirtualKitti~\cite{virtual_kitti}, and
Waymo Open~\cite{Sun_2020_CVPR}.
We train on 48-frame clips at $256{\times}256$ resolution, decoding 2048 random queries that are oversampled on specific regions for more efficient training
, see the appendix for details.
The model is trained for 500\,k steps using the AdamW optimizer~\cite{adam, adamw} with a local batch size of 1 across 64 TPU chips, taking a total of just over 2 days to complete.

\begin{table}[t]
    \centering
    \label{tab:trackable_points}
    \resizebox{0.98\linewidth}{!}{
    \small
    \begin{tabular}{@{}lrrrr}
        \toprule
        \multirow{2}[2]{*}{\textbf{Method}} & \multicolumn{4}{c}{\textbf{Max.\ Track Count @ Target FPS}} \\ 
        \cmidrule(lr){2-5} 
                                        & {60 FPS} & {24 FPS} & {10 FPS} & {1 FPS} \\ 
        \midrule
        {DELTA}~\cite{ngo2024delta} & 0 & 5 & 408 & 5,770 \\
        {SpatialTrackerV2}~\cite{xiao2025spatialtrackerv2} & 29 & 84 & 219 & 2,290 \\ 
        \midrule
        \textbf{\name (Ours)} ~~~~~~~~~~~~~ & \best{550} & \best{1,570} & \best{3,890} & \best{40,180} \\
        \bottomrule
    \end{tabular}
    }
\caption{\textbf{3D tracking throughput} --
We measure the maximum number of full-video 3D point tracks that different model can produce while maintaining a given FPS target on a single A100 GPU.
Note that for \name, each track consists of $T$ independent queries processed by the decoder. \name is 18--300$\times$ faster than others.
\vspace{-2mm}
}
\label{tab:fps}
\end{table}

\begin{table*}[t]
\centering
\small
\tabcolsep=0.5em
\resizebox{\linewidth}{!}{
\begin{tabular}{@{}cl ccc ccc ccc cc cc}
\toprule
\multirow{4}[4]{*}{\shortstack{\textbf{w/ GT} \\ \textbf{intrin.}}} & \multirow{4}[4]{*}{\textbf{Method}} & \multicolumn{13}{c}{\textbf{TAPVid-3D}} \\

\cmidrule(lr){3-15}

 & & \multicolumn{9}{c}{\textbf{Camera Coordinate 3D tracking} } & \multicolumn{4}{c}{\textbf{World Coordinate 3D tracking} }\\

\cmidrule(lr){3-11} \cmidrule(lr){12-15}
 & & \multicolumn{3}{c}{\textbf{DriveTrack}} & \multicolumn{3}{c}{\textbf{ADT}} & \multicolumn{3}{c}{\textbf{PStudio}} & \multicolumn{2}{c}{\textbf{DriveTrack}} & \multicolumn{2}{c}{\textbf{ADT}} \\ 
\cmidrule(lr){3-5}\cmidrule(lr){6-8}\cmidrule(lr){9-11}\cmidrule(lr){12-13} \cmidrule(lr){14-15} 
 & & AJ~$\uparrow$  & $\text{APD}_{3D}$~$\uparrow$  & OA~$\uparrow$
  & AJ~$\uparrow$  & $\text{APD}_{3D}$~$\uparrow$  & OA~$\uparrow$
  & AJ~$\uparrow$  & $\text{APD}_{3D}$~$\uparrow$  & OA~$\uparrow$  & $\text{APD}_{3D}$~$\uparrow$  & L1~$\downarrow$ & $\text{APD}_{3D}$~$\uparrow$  & L1~$\downarrow$\\ 
\midrule
\multirow{4}[6]{*}{\ding{55}} & St4RTrack~\cite{feng2025st4rtrack}
& - & - & -
& - & - & -
& - & - & - 
& 0.020 & 0.499 & 0.062 & 0.839 \\
& CoTracker3~\cite{karaev2025cotracker3} + UniDepthV2~\cite{unidepthv2}
& 0.038 & 0.062 & 0.856
& 0.102 & 0.146 & \best{0.937}
& \secondbesttrack{0.119} & \secondbesttrack{0.176} & \secondbesttrack{0.862} 
& - & - & - & -\\
& CoTracker3~\cite{karaev2025cotracker3} + VGGT~\cite{wang2025vggt} 
& \secondbesttrack{0.129} & \secondbesttrack{0.189} & 0.856
& 0.132 & 0.190 & \best{0.937}
& 0.045 & 0.070 & \secondbesttrack{0.862}
& \secondbesttrack{0.205} & 0.170 & 0.192 & 0.736\\
 & SpatialTrackerV2~\cite{xiao2025spatialtrackerv2} 
& 0.064 & 0.100 & \secondbesttrack{0.865} 
& \best{0.260} & \best{0.342} & 0.936 
& 0.097 & 0.152 & 0.805 
& 0.101 & \secondbesttrack{0.139} & \best{0.336} & \secondbesttrack{0.238}\\

\cmidrule(l){2-15}

 & \textbf{\name (Ours)} 
& \best{0.257} & \best{0.345} & \best{0.875}
& \secondbesttrack{0.240} & \secondbesttrack{0.319} & {0.926} 
& \best{0.138} & \best{0.186} & \best{0.897}
& \best{0.373} & \best{0.020} & \secondbesttrack{0.319} & \best{0.096} \\
\midrule\midrule
\multirow{4}[6]{*}{\ding{51}} & CoTracker3~\cite{karaev2025cotracker3} + UniDepthV2~\cite{unidepthv2} 
& 0.147 & 0.225 & 0.856
& 0.173 & 0.254 & \best{0.937}
& {0.205} & 0.311 & \secondbesttrack{0.862} 
& - & - & - & -\\
 & CoTracker3~\cite{karaev2025cotracker3} + VGGT~\cite{wang2025vggt} 
& \secondbesttrack{0.245} & \secondbesttrack{0.342} & 0.856
& 0.175 & 0.250 & \best{0.937}
& \secondbesttrack{0.215} & \secondbesttrack{0.318} & \secondbesttrack{0.862}
& \secondbesttrack{0.292} & \secondbesttrack{0.160} & 0.234 & 0.755\\
& DELTA~\cite{ngo2024delta} 
& 0.170  & 0.238  & \secondbesttrack{0.874}
& 0.199  & 0.258 & 0.879
& 0.175  & 0.261  & 0.760
& - & - & - & -\\
 & SpatialTrackerV2~\cite{xiao2025spatialtrackerv2} 
& 0.195 & 0.275 & {0.865} 
& \secondbesttrack{0.303} & \secondbesttrack{0.404} & 0.936 
& 0.175 & 0.270 & 0.805 
& 0.201 & 0.117 & \secondbesttrack{0.378} & \secondbesttrack{0.263} \\
\cmidrule(l){2-15}
 & \textbf{\name (Ours)} 
& \best{0.304} & \best{0.410} & \best{0.875}
& \best{0.307} & \best{0.408} & {0.926} 
& \best{0.372} & \best{0.498} & \best{0.897}
& \best{0.470} & \best{0.017} & \best{0.398} & \best{0.093} \\
\bottomrule
\end{tabular}
}
\caption{
    \textbf{4D reconstruction and tracking} --
    We evaluate 3D tracking capability on dynamic videos, with tracks predicted in both local camera coordinates (left) and world coordinates (right). Our model achieves superior performance compared to the prior state-of-the-art.
}
\label{tab:3d_track}
\end{table*}

\subsection{Qualitative Analysis}
\label{sec:exp:qualitative}

We begin our evaluations by inspecting video reconstruction quality in \cref{fig:point_cloud_vis}.
Dynamic objects (\eg, swan and train) reveal qualitative differences, highlighting performance gaps between different methods.
Pure reconstruction methods (see \cref{sec:related:recon}), exemplified by the state-of-the-art models MegaSaM and $\pi^3$, fail to understand dynamics in the scenes, producing either repeated images of the entities as they move through the scene, or failing to reconstruct them altogether.
Conversely, tracking-based models successfully reconstruct and track dynamics.
However, they commonly only track points from one frame, leaving gaps in the reconstruction in all areas that are occluded in the first frame as seen for SpatialTrackerV2~\cite{xiao2025spatialtrackerv2} here.
Meanwhile, \name is able to track all dynamic pixels of the video in a unified reference frame.
We further evaluate our method on diverse in-the-wild videos in \cref{fig:ours_vis}, demonstrating its performance to both static and dynamic scenes.

\subsection{\task}
\label{sec:exp:4d}

\begin{table*}[tb!]
\centering
\small
\tabcolsep=0.5em
\resizebox{\linewidth}{!}{
\begin{tabular}{@{}l cc cc cc cc cc}
\toprule
 \multirow{3}[3]{*}{\textbf{Method}} & \multicolumn{2}{c}{\textbf{Point Cloud}} & \multicolumn{8}{c}{\textbf{Video Depth}} \\
\cmidrule(lr){2-3}\cmidrule(lr){4-11} 
 & {\;\;\;\textbf{Sintel}\;\;\;} & {\textbf{Scannet}} & \multicolumn{2}{c}{\textbf{Sintel}} & \multicolumn{2}{c}{\textbf{ScanNet}}  & \multicolumn{2}{c}{\textbf{KITTI}} & \multicolumn{2}{c}{\textbf{Bonn}} \\ 
\cmidrule(lr){2-2}\cmidrule(lr){3-3}\cmidrule(lr){4-5}\cmidrule(lr){6-7}\cmidrule(lr){8-9}\cmidrule(lr){10-11}
 & L1~$\downarrow$ & L1~$\downarrow$ & AbsRel (S)~$\downarrow$ & AbsRel (SS)~$\downarrow$ & AbsRel (S)~$\downarrow$ & AbsRel (SS)~$\downarrow$ &  AbsRel (S)~$\downarrow$ & AbsRel (SS)~$\downarrow$   & AbsRel (S)~$\downarrow$ & AbsRel (SS)~$\downarrow$ \\ 
\midrule
MegaSaM~\cite{li2025megasam} &  1.531 & 0.072 &  0.342 & 0.249 & 0.050 & 0.047 & 0.109 &  0.101 & 0.056 & 0.056    \\
VGGT~\cite{wang2025vggt} & 1.582 &  0.063 & 0.318 & 0.247 & 0.044 & 0.040 & 0.094 & 0.067  & 0.055 & 0.051 \\
MapAnything~\cite{keetha2025mapanything} &  1.718  & 0.064 & 0.397 & 0.306  & 0.043 & 0.035 & 0.096 &   0.090 & 0.076 & 0.049 \\ 
SpatialTrackerV2~\cite{xiao2025spatialtrackerv2} & 1.375 & 0.036 & \secondbestdepth{0.209} & {0.175} & 0.027 & 0.025  & 0.075 & 0.064 & {0.042} & {0.978} \\
$\pi^3$~\cite{wang2025pi} & \secondbestdepth{1.139} & \secondbestdepth{0.030} & 0.241 & \secondbestdepth{0.163} & \secondbest{0.021} & \secondbest{0.019} & \best{0.055} & \secondbestdepth{0.053} & \best{0.033} & \best{0.032}   \\
\midrule
\textbf{\name (Ours)} & \best{0.768} & \best{0.028} & \best{0.171} & \best{0.148}  & \best{0.020} & \best{0.018}  & \best{0.055} & \best{0.051}  & \secondbestdepth{0.036} & \secondbestdepth{0.036}    \\
\bottomrule
\end{tabular}
}
\caption{\textbf{Video depth and point map estimation} -- Quantitative results for both video depth and point map estimation across four benchmarks. D4RT achieves top-tier performance on the depth estimation task under both scale-only (S) and scale-and-shift (SS) alignments.}
\label{tab:depth}
\end{table*}

We now turn to the quantitative evaluation of \task, comparing \name to other methods that can estimate dynamic correspondences.
We evaluate on TAPVid-3D~\cite{koppula2024tapvid} which is comprised of three subsets of real-world videos in challenging settings.

We first evaluate 3D tracking in the local camera frame using the standard protocol from \citet{koppula2024tapvid}, which measures a model's ability to predict and track the 3D position of every observed point in the \emph{local} camera reference frames.
We report standard 3D tracking metrics \emph{Average percent of points within delta error} $\text{APD}_{3D}$, \emph{Occlusion Accuracy} (OA) and \emph{3D Average Jaccard} (AJ).
As shown in \cref{tab:3d_track} (left), \name achieves state-of-the-art results against competing methods, both with and without known ground-truth intrinsics.

\cref{tab:3d_track} (right) shows results for \emph{world coordinate 3D tracking}.
This task measures a model's ability to predict tracks within a single, consistent world coordinate system, thereby also measuring the models' ability to implicitly change reference frames.
PStudio is omitted here as it has no camera motion, rendering this task effectively identical to the preceding one.
Since there is no concept of occlusion for this task, we report $\text{APD}_{3D}$ as well as the L1 distance between the predicted and ground-truth tracks which provides a comprehensive, global measure of deviation.
We observe that our model excels at this task as well, exhibiting strong improvements across metrics.

We finally compare efficiency in
\cref{tab:fps}, by counting the number of tracks each model can produce for a given target frame-per-second rate (FPS).
The measurements show that our model is 18-300$\times$ faster than prior methods.

\begin{table}[t]
\centering
\small
\tabcolsep=0.3em
\resizebox{\linewidth}{!}{
\begin{tabular}{@{}l ccc ccc c}
\toprule
\multirow{2}[2]*{\textbf{Method}} & \multicolumn{3}{c}{\textbf{Sintel}} & \multicolumn{3}{c}{\textbf{ScanNet}} & \textbf{Re10K} \\ 
\cmidrule(lr){2-4}\cmidrule(lr){5-7} \cmidrule(lr){8-8}
  & ATE~$\downarrow$& RPE-T~$\downarrow$& RPE-R~$\downarrow$& ATE~$\downarrow$& RPE-T~$\downarrow$& RPE-R~$\downarrow$ &Pose AUC~$\uparrow$ \\ 
\midrule
MegaSaM~\cite{li2025megasam} & \secondbest{0.074} & \secondbest{0.030} & \best{0.126} & 0.029 & 0.016 & 0.839 & 71.0  \\
VGGT~\cite{wang2025vggt} & 0.168 & 0.056 & 0.428 & 0.016 & 0.012  & 0.316 & 70.2 \\
MapAnything~\cite{keetha2025mapanything} & 0.202 & 0.089 & 2.383 & 0.023 & 0.016 & 0.438 & 68.7  \\
SpatialTrackerV2~\cite{xiao2025spatialtrackerv2} & 0.126 & 0.053 & 1.052 & 0.018 & 0.012 & {0.324} & 75.7 \\
$\pi^3$~\cite{wang2025pi} & 0.086 & 0.039 & 0.248 & \secondbest{0.015} & \best{0.010} & \best{0.291} & \secondbest{78.7}  \\
\midrule
\textbf{\name (Ours)} & \best{0.065} & \best{0.024} & \best{0.126} & \best{0.014} & \best{0.010} & \secondbest{0.302} & \best{83.5} \\
\bottomrule
\end{tabular}
}

\caption{\textbf{Camera pose estimation} -- We evaluate D4RT against state-of-the-art methods on static indoor scenes (ScanNet, Re10K) and dynamic outdoor scenes (Sintel).}
\label{tab:camera}
\end{table}

\subsection{3D Reconstruction}
\label{sec:exp:3d}

We now evaluate the 3D reconstruction capabilities of our model.
The most holistic task here is 3D point cloud reconstruction, which requires predicting all observed pixels in the same world coordinate system.

We evaluate on two standard benchmarks: MPI Sintel~\cite{bulter2012sintel} and ScanNet~\cite{scannet}, which consist of dynamic and static scenes, respectively.
For evaluation, we first align the predicted and ground-truth point clouds via mean-shifting following \citet{li2025megasam} before reporting the mean L1 distance.
As shown in \cref{tab:depth} (left), our model outperforms all recent state-of-the-art models across datasets.

\para{Depth estimation.}
We evaluate depth estimation performance across four datasets: Sintel~\cite{bulter2012sintel},
ScanNet~\cite{scannet},
KITTI~\cite{kitti2013}, and
Bonn~\cite{Bonn2019}.
Following prior works~\cite{li2025megasam, wang2025cut3r}, we align the predicted video depth with the ground-truth depth before computing metrics.
Specifically, we adopt two alignment settings:
\emph{scale-only} (S) and \emph{scale-and-shift} (SS) alignment.
Both settings determine a single global scale factor between the predicted and ground-truth depths across the sequence, while the latter introduces an additional 3D translation term, following an affine-invariant setup.
After alignment, we compute the \emph{Absolute Relative Error} (AbsRel)~\cite{AbsRelErr} between the predicted and ground-truth depths.

We show results in \cref{tab:depth} (right).
\name achieves top-tier performance across all benchmarks, with particularly strong results on Sintel which is the most challenging dataset due to its highly dynamic scenes.

\para{Camera pose estimation.}
We evaluate camera pose estimation performance on the same datasets.
Following prior work~\cite{wang2025cut3r, jiang2025geo4d}, we evaluate on the $14$-sequence subset of Sintel which provides full rendering with motion blur and atmospheric effects to measure in-the-wild performance.
We report the \emph{Absolute Translation Error} (ATE), \emph{Relative Translation Error} (RPE trans), and \emph{Relative Rotation Error} (RPE rot) after Sim(3) alignment with the ground truth as in \cite{zhang2024monst3r, wang2025cut3r, wang2025pi}, as well as the \emph{Pose AUC@$30$} as in \cite{wang2025vggt}.
\cref{tab:camera} presents the results, where our model strongly outperforms competing baselines on both indoor scenes (ScanNet and Re10K) and simulated outdoor scenes (Sintel).

\cref{fig:speed} visualizes efficiency and accuracy for camera pose estimation across several recent state-of-the-art methods.
To ensure a fair comparison, we remove the decoding heads of the baselines which are unrelated to camera estimation, making them faster.
Despite this, we find that \name surpasses all existing methods in both accuracy and efficiency, notably outperforming MegaSaM in throughput by two orders of magnitude.

\subsection{Ablation Studies}
\label{sec:exp:ablations}

We present a series of ablation studies to validate our design choices and hyperparameters.
Unless otherwise noted, we adopt a ViT-L backbone with $6$ decoder layers as the default setting.
The model is trained for 300\,k iterations with a batch size 16.
All results are reported on the challenging Sintel dataset for both video depth and camera pose.

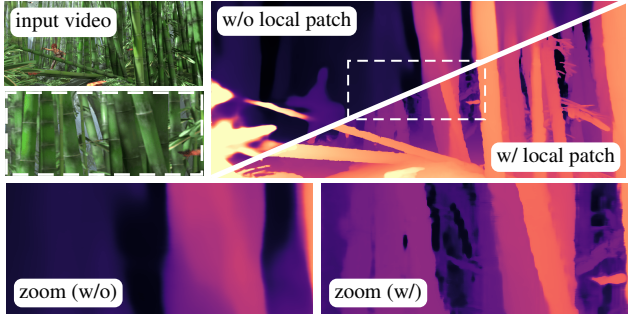
\begin{figure}[t]
    \newcommand{\imagelabel}[1]{%
        \tikz\node[fill=white, rounded corners=3pt, inner sep=2pt, font=\footnotesize, text=black]{#1};%
    }

    \newcommand{\sideImage}[2]{%
    \begin{tikzpicture}[baseline=(zoomImage.south)]
        \pgfmathsetlengthmacro{\stackwidth}{0.333\linewidth-4pt}
    
        \node[anchor=south west, inner sep=0] (zoomImage) at (0,0) {
            \includegraphics[width=\stackwidth, viewport=358 152 665 283, clip]{#1}
        };
        \draw [white, dash pattern=on 6pt off 3.6pt, line width=1pt] ($(zoomImage.south west) + (0.7pt, 0.7pt)$) rectangle ($(zoomImage.north east) - (0.7pt, 0.7pt)$);
    
        \node[anchor=south west, inner sep=0, outer sep=0] (mainImage) at ([yshift=2pt]zoomImage.north west) {
            \includegraphics[width=\stackwidth]{#1}
        };
    
        \node[anchor=north west, inner sep=0] at ([xshift=2pt, yshift=-2pt]mainImage.north west) {\imagelabel{#2}};
    \end{tikzpicture}}

    \newcommand{\diagCollage}[2]{%
    \begin{tikzpicture}
        \pgfmathsetlengthmacro{\imgwidth}{0.6666\linewidth-1.5pt} 
        
        \node[anchor=south west, inner sep=0] (image) at (0,0) {\includegraphics[width=\imgwidth]{#2}};
        \path (image.south west) coordinate (SW) (image.north east) coordinate (NE);
        
        \begin{scope}
            \clip (SW) -- (image.north west) -- (NE) -- cycle;
            \node[anchor=south west, inner sep=0] at (0,0) {\includegraphics[width=\imgwidth]{#1}};
        \end{scope}
        \draw[white, line width=2.0pt] (SW) -- (NE);

        \draw [white, dash pattern=on 4pt off 2.4pt, line width=0.75pt] 
              ($(image.south west)!0.33!(image.north east)$) rectangle ($(image.south west)!0.66!(image.north east)$);

        \node[anchor=north west, inner sep=0] at ([xshift=2pt, yshift=-2pt]image.north west) {\imagelabel{w/o local patch}};
        \node[anchor=south east, inner sep=0] at ([xshift=-2pt, yshift=2pt]image.south east) {\imagelabel{w/ local patch}};
    \end{tikzpicture}}

    \newcommand{\zoomPair}[2]{%
    \begin{tikzpicture}
        \pgfmathsetmacro{\zoomwidth}{(\linewidth-5.5pt)/2}

        \node[anchor=south west, inner sep=0] (z1) at (0,0) {
            \includegraphics[width=\zoomwidth pt, viewport=358 152 665 283, clip]{#1}
        };
        \node[anchor=south west, inner sep=0] at ([xshift=3pt, yshift=2.5pt]z1.south west) {\imagelabel{zoom (w/o)}};

        \node[anchor=south west, inner sep=0] (z2) at ([xshift=3pt]z1.south east) {
            \includegraphics[width=\zoomwidth pt, viewport=358 152 665 283, clip]{#2}
        };
        \node[anchor=south west, inner sep=0] at ([xshift=3pt, yshift=2.5pt]z2.south west) {\imagelabel{zoom (w/)}};
    \end{tikzpicture}}
    
    \setlength\tabcolsep{1pt} 
    \renewcommand{\arraystretch}{0.35}
    \small \centering
    \usetikzlibrary{calc} %

    \begin{tabular}{@{}cc@{}} 
        \raisebox{1.5pt}{\sideImage{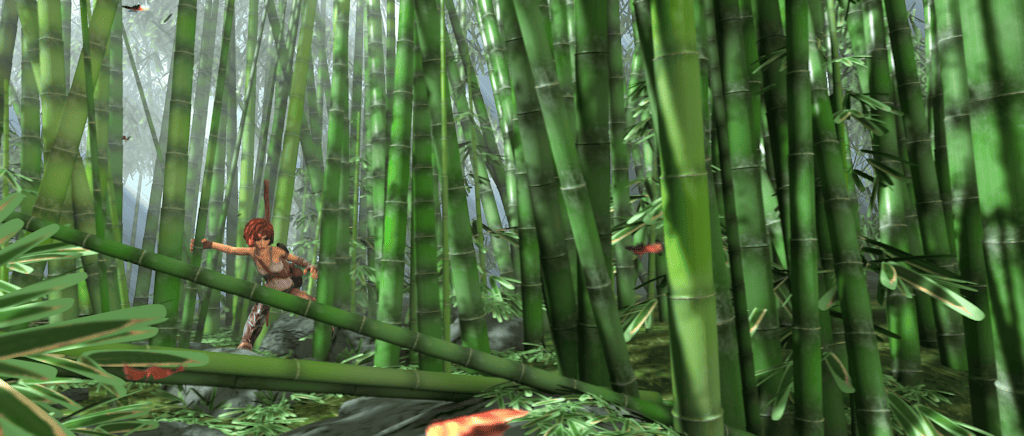}{input video}} & 
        \diagCollage{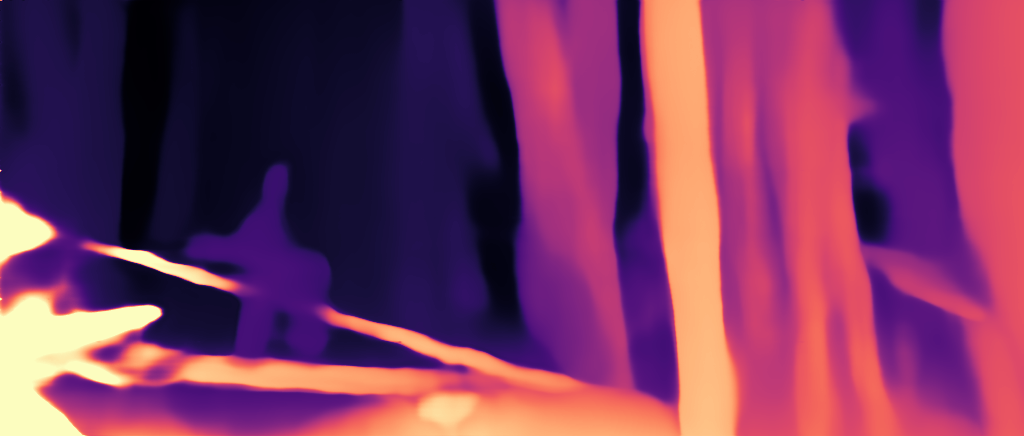}{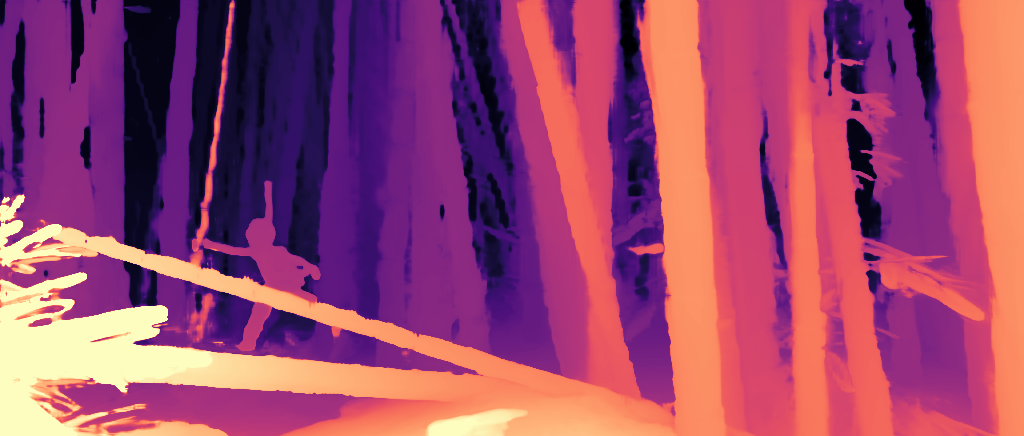} \\[-1pt]
        
    \end{tabular}
    \begin{tabular}{@{}c@{}} 
    \zoomPair{imgs/patch/np_5.png}{imgs/patch/lp_5.png}
    \end{tabular}
    
    \caption{\textbf{Preservation of low-level details} -- We ablate the effect of including local patch information into the model's queries, visualizing the resulting depth maps on an example from the Sintel dataset. We find that the local RGB patches help preserve fine-grained details and produce sharper object boundaries.}
    \label{fig:ablation_source_patch}
\end{figure}

\begin{table}[tb!]
\centering
\small
\tabcolsep=0.5em
\resizebox{\linewidth}{!}{
\begin{tabular}{@{}c cc ccc}
\toprule
\multirow{2}[2]{*}{\textbf{\shortstack{w/ local \\ appear. patch}}} & \multicolumn{2}{c}{\textbf{Video Depth}} & \multicolumn{3}{c}{\textbf{Camera Pose}} \\ 
\cmidrule(lr){2-3}\cmidrule(lr){4-6}
  & AbsRel (S)~$\downarrow$  & AbsRel (SS)~$\downarrow$  & ATE~$\downarrow$ & RPE-T~$\downarrow$& RPE-R~$\downarrow$ \\ 
\midrule
\ding{55}  & 0.366 & 0.306 & 0.173 & 0.031 & 0.262 \\
\ding{51}  & \best{0.302} & \best{0.257} & \best{0.091} & \best{0.028} & \best{0.245} \\
\bottomrule
\end{tabular}
}

\caption{\textbf{Local RGB patch} -- We evaluate a ViT-L model, trained with and without the appearance patch as input to the decoder, on Sintel video depth and camera pose estimation.}
\label{tab:abla_source_patch}
\end{table}

\para{Local RGB patch.}
As described in \cref{sec:method:architecture}, we introduce an additional embedding that encodes the local appearance around the query's source location.
This additional embedding yields dramatic performance improvements as shown in \cref{tab:abla_source_patch} and \cref{fig:ablation_source_patch}. These improvements can be attributed to two factors: (i) the local appearance information helps the query establish more reliable correspondences with the encoded spatiotemporal features; and (ii) these patches provide low-level cues that help to segment objects from their surroundings, leading to fine-grained predictions, as exemplified by the sharper depth estimates. We note that while most related works have specialized modules to preserve low level details, such as DPT~\cite{Ranftl2021} with its skip connections between encoder and decoder layers, our approach is much simpler.
In \cref{sec:app:high-res} of the appendix, we further highlight how the patches unlock subpixel precision.

\para{Encoder size.}
We now examine how performance scales with the size of our encoder.
\cref{tab:abla_model_size} summarizes our findings for ViT-B (90\,M) to ViT-g (1\,B).
As the number of trainable parameters increases, we observe a significant performance improvement, particularly in depth estimation and RPE-R in camera pose estimation.

\begin{table}[t]
\centering
\small
\tabcolsep=0.5em
\resizebox{\linewidth}{!}{
\begin{tabular}{@{}l cc ccc}
\toprule
\multirow{2}[2]{*}{\textbf{Losses}}& \multicolumn{2}{c}{\textbf{Video Depth}} & \multicolumn{3}{c}{\textbf{Camera Pose}} \\
\cmidrule(lr){2-3}\cmidrule(lr){4-6}
&AbsRel(S)~$\downarrow$&AbsRel(SS)~$\downarrow$&ATE~$\downarrow$&RPE-T~$\downarrow$&RPE-R~$\downarrow$\\
\midrule
\name&0.302&0.257&0.091&0.028&0.245\\
\midrule
w/o 2D position&\cellcolor[rgb]{1.000,0.500,0.500}{+0.071}&\cellcolor[rgb]{1.000,0.500,0.500}{+0.063}&\cellcolor[rgb]{1.000,0.937,0.937}{+0.002}&\cellcolor[rgb]{1.000,0.909,0.909}{+0.002}&\cellcolor[rgb]{0.878,1.000,0.878}{-0.028}
\\
w/o normal&\cellcolor[rgb]{1.000,0.611,0.611}{+0.043}&\cellcolor[rgb]{1.000,0.679,0.679}{+0.026}&\cellcolor[rgb]{1.000,0.923,0.923}{+0.003}&\cellcolor[rgb]{1.000,0.889,0.889}{+0.003}&\cellcolor[rgb]{0.904,1.000,0.904}{-0.022}
\\
w/o displacement&\cellcolor[rgb]{1.000,0.803,0.803}{+0.011}&\cellcolor[rgb]{1.000,0.833,0.833}{+0.007}&\cellcolor[rgb]{1.000,0.852,0.852}{+0.011}&\cellcolor[rgb]{1.000,0.889,0.889}{+0.003}&\cellcolor[rgb]{1.000,0.877,0.877}{+0.007}
\\
w/o visibility&\cellcolor[rgb]{0.979,1.000,0.979}{-0.003}&\cellcolor[rgb]{0.802,1.000,0.802}{-0.025}&\cellcolor[rgb]{1.000,0.846,0.846}{+0.012}&\cellcolor[rgb]{1.000,0.788,0.788}{+0.011}&\cellcolor[rgb]{1.000,0.781,0.781}{+0.022}
\\
w/o confidence&\cellcolor[rgb]{1.000,0.916,0.916}{+0.002}&\cellcolor[rgb]{0.802,1.000,0.802}{-0.025}&\cellcolor[rgb]{1.000,0.500,0.500}{+0.126}&\cellcolor[rgb]{1.000,0.500,0.500}{+0.061}&\cellcolor[rgb]{1.000,0.500,0.500}{+0.115}
\\
\bottomrule
\end{tabular}
}
\caption{\textbf{Auxiliary losses} --
We remove auxiliary losses individually to evaluate their impact on model performance.
A slight trade-off between depth and camera estimation pose is observed, though we find that all auxiliary losses improve overall performance.
}
\label{tab:ablation_loss}
\end{table}

\begin{table}[tb!]
\centering
\small
\tabcolsep=0.5em
\resizebox{\linewidth}{!}{
\begin{tabular}{@{}c cc ccc}
\toprule
\multirow{2}[2]{*}{\textbf{\shortstack{Backbone \\ size}}} & \multicolumn{2}{c}{\textbf{Video Depth}} & \multicolumn{3}{c}{\textbf{Camera Pose}} \\ 
\cmidrule(lr){2-3}\cmidrule(lr){4-6}
  & AbsRel (S)~$\downarrow$  & AbsRel (SS)~$\downarrow$ & ATE~$\downarrow$ & RPE-T~$\downarrow$& RPE-R~$\downarrow$ \\ 
\midrule
ViT-B  & 0.319 & 0.232 & 0.145 & 0.034 & 0.266 \\
ViT-L  & 0.256 & 0.214 & {0.073} & \secondbest{0.027} & 0.191 \\
ViT-H  & \secondbest{0.226} & \secondbest{0.173} & \best{0.070} & 0.028 & \secondbest{0.186} \\
ViT-g  & \best{0.191} & \best{0.168} & \secondbest{0.078}  &  \best{0.026} & \best{0.160} \\
\bottomrule
\end{tabular}
}

\caption{\textbf{Backbone size} -- We examine how D4RT's performance scales with the size of the pretrained ViT encoder backbone, evaluating video depth and camera pose estimation performance on Sintel. We observe a clear improvement as the backbone size increases from ViT-B to ViT-g.}
\label{tab:abla_model_size}
\end{table}

\para{Auxiliary losses.}
We finally investigate the contribution of each auxiliary loss in \cref{tab:ablation_loss}.
We observe a range of behaviors acros the auxiliary losses:
some significantly boost depth performance (\eg, 2D position, normal), others are critical for better camera pose estimation (\eg, confidence), and displacement has small improvements across the board.

\section{Conclusion}
\label{sec:conclusion}

In this work, we introduce \name, a simple yet highly scalable feedforward network that reconstructs dynamic 4D scenes with temporal correspondence.
Our core innovation is an efficient, query-based decoder that allows for the independent prediction of any point's 3D position in space and time.
This flexible parametrization avoids the computational bottlenecks of dense per-frame decoders, enabling inference that scales linearly with the number of points to be reconstructed.
Crucially, we demonstrate that \name achieves state-of-the-art results across a wide range of 4D tasks including depth,
point cloud estimation, and 3D point tracking.
\name demonstrates that scaling to complex, dynamic environments does not require sacrificing precision, offering a unified framework for the next generation of 4D perception.

\para{Contribution and Acknowledgements.}
MS led the project, with management support from JZ. MS proposed the SRT-style decoder, and GL proposed local RGB patches and tracking-all-pixels. CZ, GL, IR, and MS contributed to the core model design. LM, SS, IR, and SK designed and created training datasets. CZ, GL, and MS carried out the major implementation, with significant contributions from SK, IR, LM, and JX. CZ drove the model experimentation. Comprehensive evaluations and data pipelines were set up by CZ, GL, SK, IR, LM, JX, SS, and MS. Visualizations were produced by GL, IR and JX. RS, JB, RH, ZG, and AZ provided project support, advising, and guidance.

\noindent We thank a number of colleagues and advisors for making this work possible. We thank Saurabh Saxena, Kaiming He, Carl Doersch, Leonidas Guibas, Noah Snavely, Ben Poole, Joao Carreira, Pauline Luc, Yi Yang,  Howard Huang, Huizhong Chen, Cordelia Schmid for providing advice during the project; Gabriel Brostow, for advising SK during the course of the project and for providing feedback on the manuscript; Relja \mbox{Aran\dj elović} and Maks Ovsjanikov for providing feedback on the manuscript; Ross Goroshin, Tengda Han and Dilara Gokay for their help during early-stage development; Aravindh Mahendran for helping with code reviews; Daniel Duckworth for helping with visualizations and comparisons against baselines; and Alberto García and Jesús Pérez for the invaluable contributions to data generation and collection.

{
    \small
    \bibliographystyle{ieeenat_fullname}
    \bibliography{main}
}

\clearpage
\appendix

\onecolumn
{
\centering
\Large
\textbf{\thetitle}\\
\vspace{0.5em}Appendix \\
\vspace{1.0em}
}

\section{Model Overview \& Training Details}
\label{sec:app:overview}
\begin{figure*}[h]
\centering
\vspace{-4mm}
\includegraphics[width=1.0\linewidth]{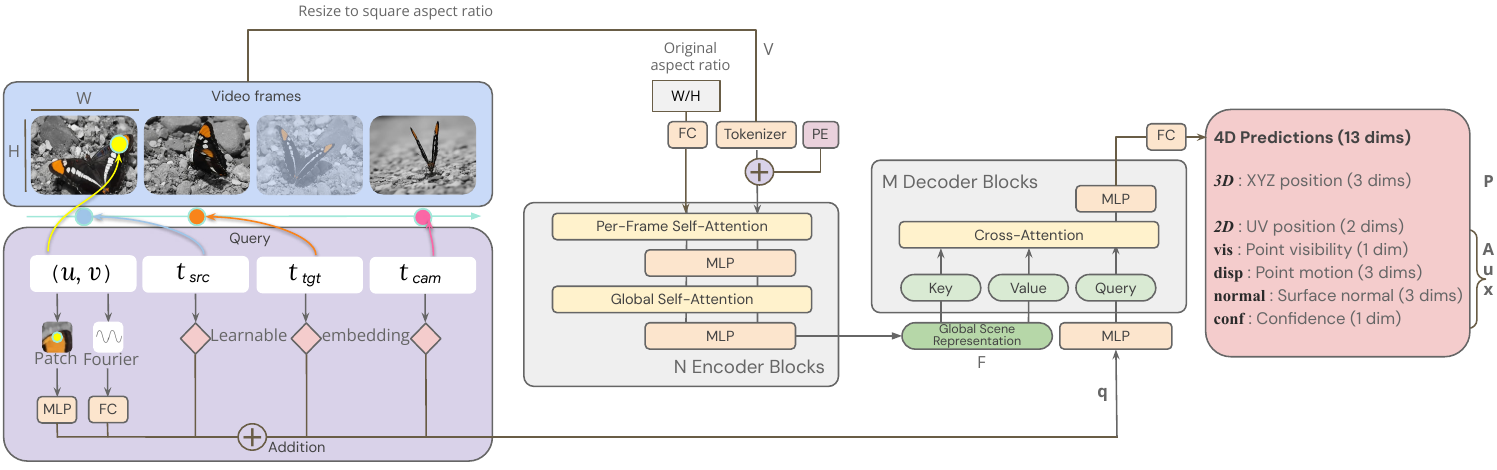}
\caption{
\textbf{Full \name model overview} --
We provide a holistic overview of the model together with its inputs and outputs.
FC corresponds to a fully connected layer, and PE for positional encoding.
See \cref{sec:method} of the main paper for reference. 
}
\label{fig:app:method}
\end{figure*}

\noindent The model is trained end-to-end by minimizing a composite loss $\mathcal{L}$, which is a weighted sum of task-specific losses computed over a batch of $N$ sampled queries:
\begin{equation*}
\mathcal{L} = \frac{1}{N} \sum_{i=1}^{N} \Big( 
    c\lambda_{3D}\mathcal{L}_{3D} 
    - \lambda_{\text{conf}}\log{c}
    + \lambda_{2D}\mathcal{L}_{2D}
    + \lambda_{\text{vis}}\mathcal{L}_{\text{vis}} 
    + \lambda_{\text{disp}}\mathcal{L}_{\text{disp}} 
    + \lambda_{\text{conf}}\mathcal{L}_{\text{conf}} 
    + \lambda_{\text{normal}}\mathcal{L}_{\text{normal}}
\Big)_i
\end{equation*}
where $c$ is the confidence score predicted by the model.
We use the following loss weights:
$\lambda_{\scriptscriptstyle \mathrm{3D}}{=}1.0$, $\lambda_{\scriptscriptstyle \mathrm{2D}}{=}0.1$, $\lambda_{\text{vis}}{=}0.1$, $\lambda_{\text{disp}}{=}0.1$,  $\lambda_{\text{conf}}{=}0.2$, $\lambda_{\text{normal}}{=}0.5$,  $\lambda_{\text{conf}}{=}0.2$.
We train using the AdamW optimizer with a weight decay of $0.03$.
The learning rate is warmed up for $2{,}500$ steps until it reaches a peak value of $10^{-4}$. 
Subsequently, it follows a cosine annealing schedule, decaying to a final value of $10^{-6}$. 
Gradients are clipped to a maximum ${L}^2$-norm of $10$.

\paragraph{Data augmentation.}
Extensive data augmentation techniques are applied to the video during training to improve model generalization.
We apply temporally consistent color jittering by performing random brightness, saturation, contrast, and hue adjustments. We also apply random color drop with a probability of $0.2$, and Gaussian blur augmentation with a probability of $0.4$. 
For spatial augmentation, we use random crop augmentations with a scale ratio between $0.3$ and $1.0$ of the original size. 
After determining the crop size, a random aspect ratio is selected by sampling from a uniform distribution in the logarithmic domain; this ensures equal probability for wide and tall crops while respecting image boundaries. 
Additionally, during the random crop augmentation, the image is randomly zoomed in with a probability of $0.05$.
On the temporal dimension, frames are subsampled from the full video with a random stride. 

\paragraph{Training queries.}
Queries are sampled from available ground truth point trajectories.
To focus the model on challenging regions, $30\%$ of the queries (along dimensions $u,v$) are sampled near depth discontinuities or motion boundaries, which are pre-computed using a Sobel filter on depth maps.
Timesteps $t_{src}$, $t_{tgt}$, and $t_{cam}$ are sampled uniformly at random, except that we enforce $t_{tgt} = t_{cam}$ with probability $0.4$ to improve downstream performance.

\clearpage

\section{Generalization to Long Videos}\label{sec:app:long-seq}
We implement a long-sequence processing algorithm by partitioning videos into overlapping segments. We then align these segments by estimating Sim(3) transformations using the Umeyama algorithm~\cite{umeyama1991least}, based on the top 85\% of points with the highest confidence in the overlapping regions. The process is similar to the first stage proposed in VGGT-Long~\cite{deng2025vggtlong}, we omit loop detection and optimization stage in ~\cite{deng2025vggtlong} to directly evaluate the reconstruction model's raw precision. As shown in \cref{fig:kitti_eval} on KITTI~\cite{kitti2013}, our model yields the best ATE in the first example, significantly outperforming VGG-T and $\pi^3$ by a large margin. On the second example, we significantly outperform VGG-T and produce a result on par with $\pi^3$. 

\begin{figure}[h]
    \centering
    \begin{subfigure}{\textwidth}
    \centering
    \includegraphics[width=0.8\linewidth]{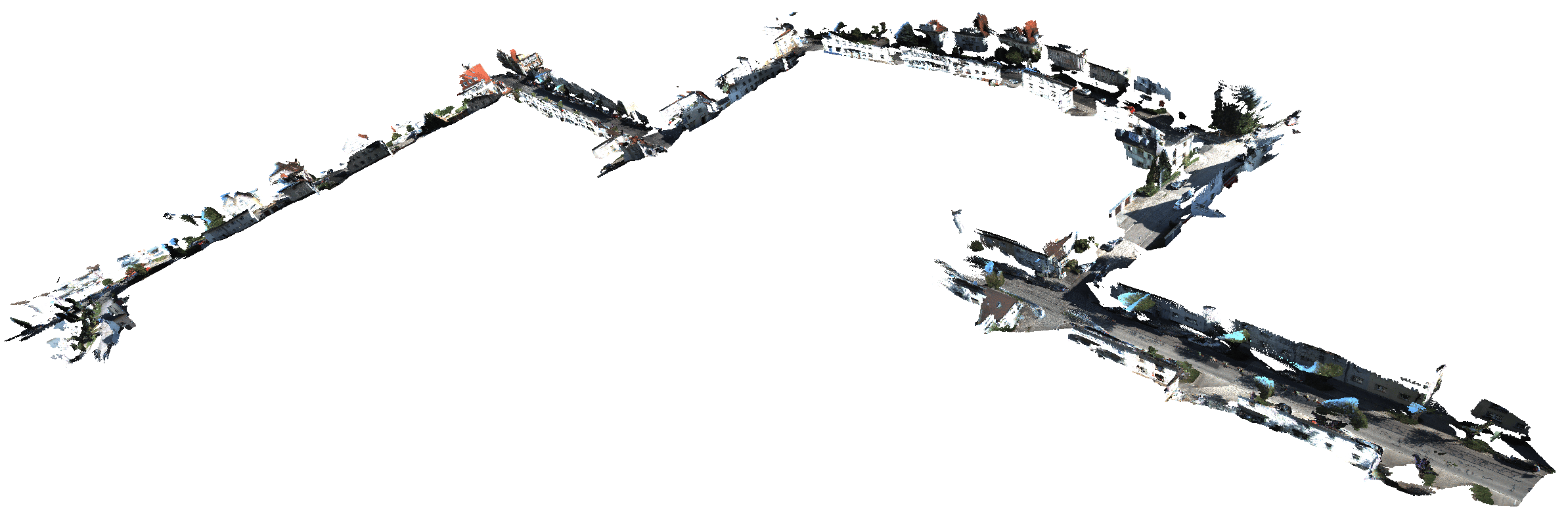}    
    \subcaption{Visualization of reconstruction of 1000 frames from KITTI sequence 00.}
    \end{subfigure}

    \begin{subfigure}{\textwidth}
    \centering
    \includegraphics[width=0.8\linewidth]{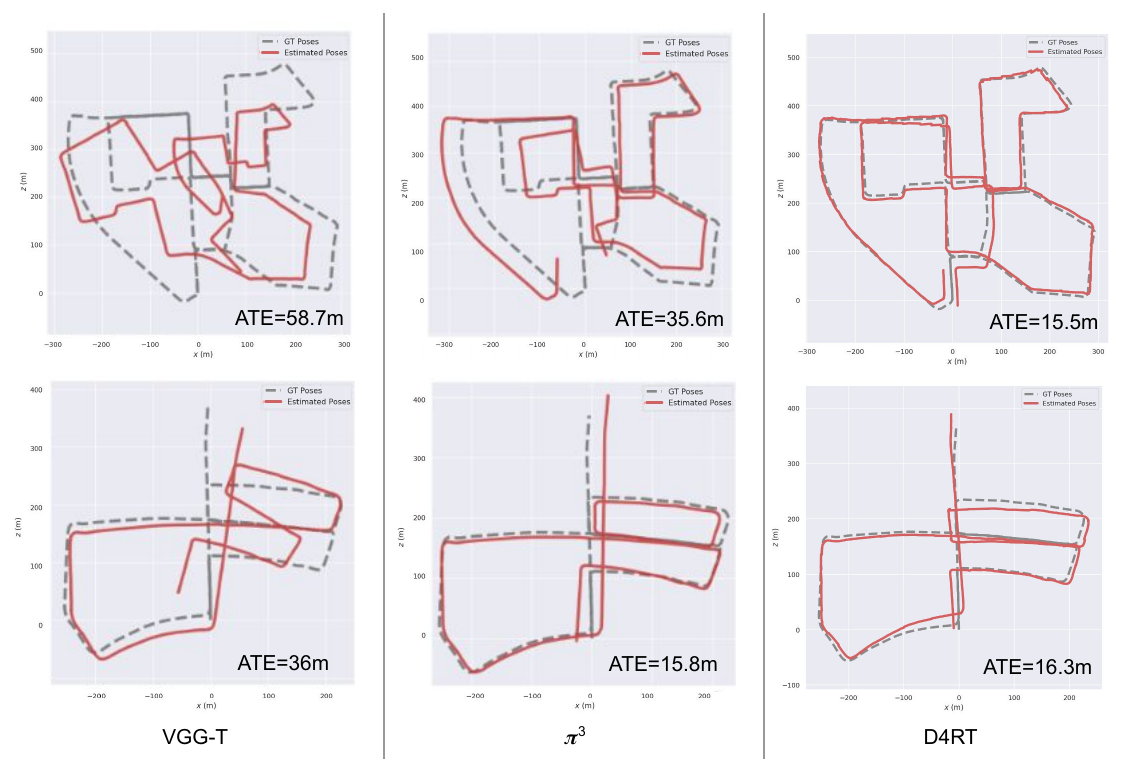}    
    \subcaption{Comparison of raw chunked prediction alignment results against VGG-T and $\pi^3$ baselines (no loop closure).}
    \end{subfigure}
    
    \caption{\textbf{Long sequence results on KITTI sequences} -- We align and stitch predictions from overlapped chunks by Umeyama algorithm, without the loop detection and global optimization proposed in~\cite{deng2025vggtlong}. Our model obtains consistently better results than VGG-T, and significantly better results than $\pi^3$ on sequence ``00" (top row).}
    \label{fig:kitti_eval}
\end{figure}

\clearpage

\section{High-Resolution Decoding with Subpixel Precision}\label{sec:app:high-res}

\noindent A key advantage of \name's architecture is the decoupling of the global scene encoding from the point-wise decoding.
As the query coordinates $(u, v)$ are defined in a continuous normalized space $[0, 1]^2$, our decoder is able to probe the scene at arbitrary resolutions, independent of the resolution of the Global Scene Representation $F$.

We explore this capability in \cref{tab:ablation_resolution}.
To quantify the preservation of high-frequency details, we report the Pseudo Depth Boundary Error accuracy ($\epsilon^{\text{acc}}_{\text{PDBE}}$) proposed by~\citet{pham2025sharpdepth}, in addition to standard depth metrics. We compare four configurations using a ViT-g encoder fixed at a resolution of $256{\times}256$.
The baseline output at the encoder's native resolution (\textbf{Config \raisebox{0.75pt}{\textcircled{\scriptsize 1}}}) is naturally coarse. Incorporating the local appearance patch mechanism described in the main paper (\textbf{Config \raisebox{0.75pt}{\textcircled{\scriptsize 2}}}) allows for the recovery of finer low-level structures, although pixelation artifacts persist.
We further leverage the continuous nature of our query mechanism to predict at the \textit{original} resolution.
While naive dense querying (\textbf{Config \raisebox{0.75pt}{\textcircled{\scriptsize 3}}}) recovers smoother edges, it still fails at recovering high frequencies.

In \textbf{Config \raisebox{0.75pt}{\textcircled{\scriptsize 4}}}, we extract the local RGB patches from the source frames at their \textit{original} resolution for decoding.

The significant drop in $\epsilon^{\text{acc}}_{\text{PDBE}}$ demonstrates how this allows \name to recover finer details.
We show qualitative results in \cref{supfig:subpixel_dec} where we observe that hair strands and object boundaries are resolved more accurately.

\begin{table}[tbh!]
\centering
\small
\tabcolsep=0.8em
\newcommand*\circled[1]{\tikz[baseline=(char.base)]{\node[shape=circle,draw,inner sep=2pt] (char) {#1};}}

\begin{tabular}{@{}cccccccccc@{}}
\toprule
\multirow{2}[2]{*}{{\textbf{Config.}}} & \multirow{2}[2]{*}{\shortstack{\textbf{Encoder} \\ \textbf{Resolution}}} & \multirow{2}[2]{*}{\shortstack{\textbf{RGB} \\ \textbf{Patch}}} & \multirow{2}[2]{*}{\shortstack{\textbf{Output} \\ \textbf{Resolution}}} & \multirow{2}[2]{*}{\shortstack{\textbf{RGB patch} \\ \textbf{Resolution}}} & \multicolumn{2}{c}{\textbf{Scale}} & \multicolumn{2}{c}{\textbf{Scale and Shift}} \\
\cmidrule(lr){6-7} \cmidrule(lr){8-9}
 & & & & & AbsRel~$\downarrow$ & $\epsilon^{\text{acc}}_{\text{PDBE}}$~$\downarrow$  & AbsRel~$\downarrow$  & $\epsilon^{\text{acc}}_{\text{PDBE}}$~$\downarrow$ \\
\midrule
\textbf{\raisebox{0.75pt}{\textcircled{\scriptsize 1}}} & $256{\times}256$ & \xmark & $256{\times}256$ & $256{\times}256$ & 0.254 & 3.323 & 0.219 & 3.307 \\
\textbf{\raisebox{0.75pt}{\textcircled{\scriptsize 2}}} & $256{\times}256$ & \cmark & $256{\times}256$ & $256{\times}256$ & 0.218 & 2.254 & 0.179 & 2.243\\
\textbf{\raisebox{0.75pt}{\textcircled{\scriptsize 3}}} & $256{\times}256$ & \cmark & \emph{Original} & $256{\times}256$ & \best{0.217} & 2.266 & {0.178} & 2.258 \\
\textbf{\raisebox{0.75pt}{\textcircled{\scriptsize 4}}} & $256{\times}256$ & \cmark & \emph{Original} & \emph{Original} & 0.220 & \best{2.193}  & \best{0.176} & \best{2.185}  \\
\bottomrule
\end{tabular}

\caption{
\textbf{Quantitative impact of query density and patch fidelity} --
Feeding RGB patches from the high-resolution video into the decoder (Config \raisebox{0.75pt}{\textcircled{\scriptsize 4}}) yields significantly sharper edges in depth maps as measured by $\epsilon^{\text{acc}}_{\text{PDBE}}$.
}

\label{tab:ablation_resolution}
\end{table}

\begin{figure*}[h]
\centering
\usetikzlibrary{calc} 

\newcommand{\labeledImage}[2]{%
    \begin{tikzpicture}
        \node[anchor=south west, inner sep=0] (image) at (0,0) {
            \includegraphics[width=0.163\linewidth]{#2}
        };
        \node[anchor=north west, fill=white, rounded corners=3pt, inner sep=2pt, font=\footnotesize, text=black] 
            at ([xshift=1pt, yshift=-0.8pt]image.north west) {#1};
    \end{tikzpicture}%
}

\makeatletter
\providecommand{\empty}{}%
\makeatother

\newcommand{\drawZoomBox}[5]{%
    \begin{tikzpicture}
        \node[anchor=south west, inner sep=0] (image) at (0,0) {
            \includegraphics[width=0.163\linewidth]{#1}
        };
        
        \begin{scope}[x={(image.south east)}, y={(image.north west)}]
            \pgfmathsetmacro{\cenX}{1 - #3}
            \pgfmathsetmacro{\cenY}{#2}
            \pgfmathsetmacro{\halfS}{#4 / 2}
            \draw[white, dash pattern=on 3pt off 1.5pt, line width=1pt] 
                (\cenX - \halfS, \cenY - \halfS) rectangle (\cenX + \halfS, \cenY + \halfS);
        \end{scope}

        \def\temp{#5}%
        \ifx\temp\empty
        \else
            \node[anchor=north west, fill=white, rounded corners=3pt, inner sep=2pt, font=\footnotesize, text=black] 
                at ([xshift=1pt, yshift=-0.8pt]image.north west) {#5};
        \fi
    \end{tikzpicture}%
}

\newcommand{\showZoomCrop}[5]{%
    \begin{tikzpicture}
        \node[anchor=south west, inner sep=0, opacity=0] (frame) at (0,0) {
            \includegraphics[width=0.163\linewidth]{#1}
        };
        
        \clip (frame.south west) rectangle (frame.north east);
        
        \begin{scope}[x={(frame.south east)}, y={(frame.north west)}]
            \pgfmathsetmacro{\scaleFactor}{1 / #4}
            \pgfmathsetmacro{\posX}{0.5 - (1 - #3) * \scaleFactor}
            \pgfmathsetmacro{\posY}{0.5 - (#2) * \scaleFactor}
            
            \node[anchor=south west, inner sep=0, scale=\scaleFactor] at (\posX, \posY) {
                \includegraphics[width=0.163\linewidth]{#1}
            };
        \end{scope}

        \def\temp{#5}%
        \ifx\temp\empty
        \else
            \node[anchor=north west, fill=white, rounded corners=3pt, inner sep=2pt, font=\footnotesize, text=black] 
                at ([xshift=1pt, yshift=-0.8pt]frame.north west) {#5};
        \fi
    \end{tikzpicture}%
}

\setlength{\tabcolsep}{1pt}
\renewcommand{\arraystretch}{0.5} %

\begin{tabular}{@{}cccccc@{}}

    \drawZoomBox{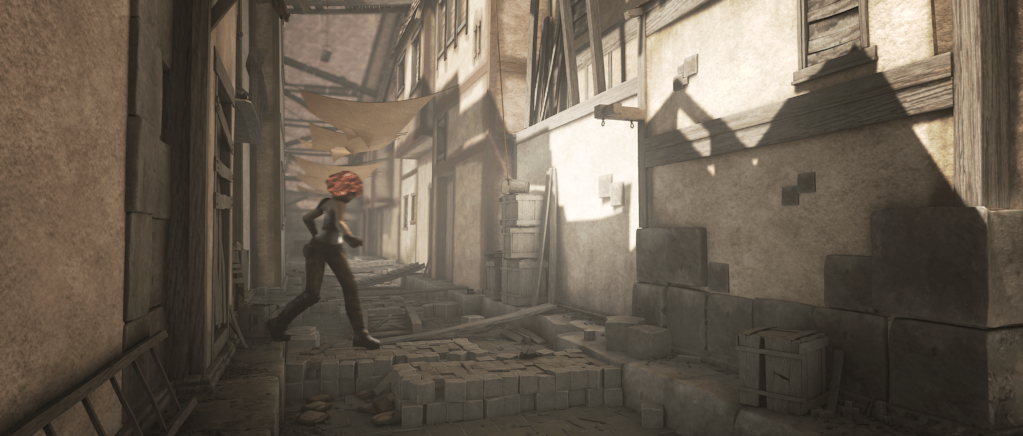}{0.35}{0.7}{0.15}{Input video} &
    \drawZoomBox{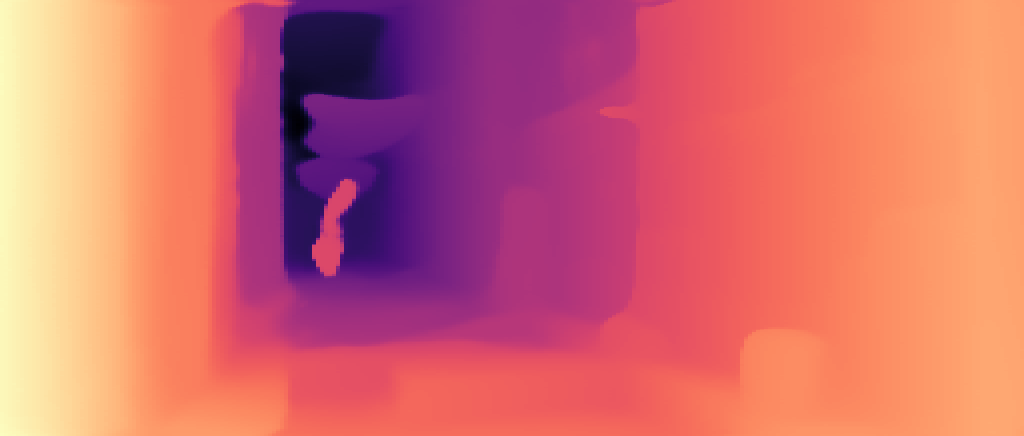}{0.35}{0.7}{0.15}{Config \textbf{{\textcircled{\scriptsize 1}}}} &
    \drawZoomBox{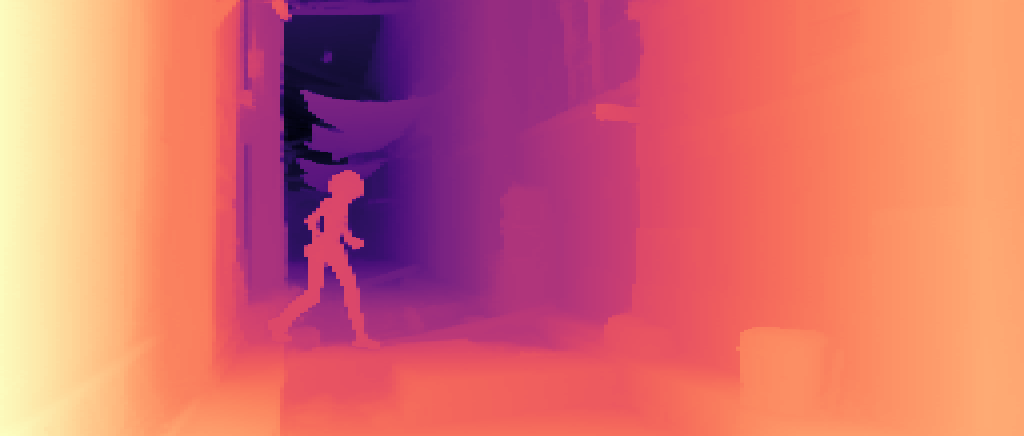}{0.35}{0.7}{0.15}{Config \textbf{{\textcircled{\scriptsize 2}}}} &
    \drawZoomBox{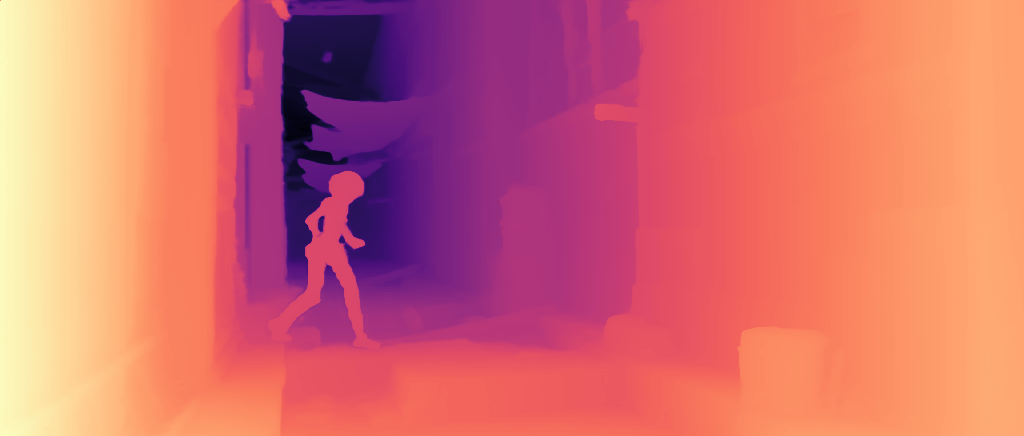}{0.35}{0.7}{0.15}{Config \textbf{{\textcircled{\scriptsize 3}}}} &
    \drawZoomBox{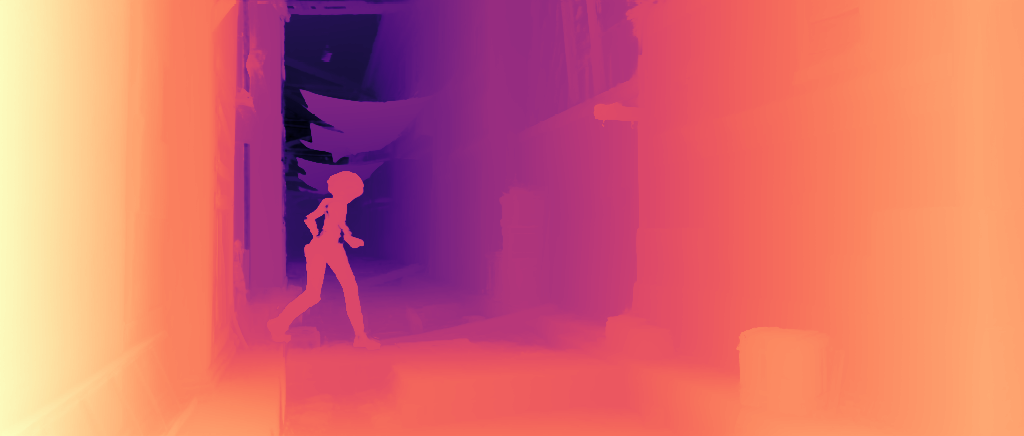}{0.35}{0.7}{0.15}{Config \textbf{{\textcircled{\scriptsize 4}}}} &
    \drawZoomBox{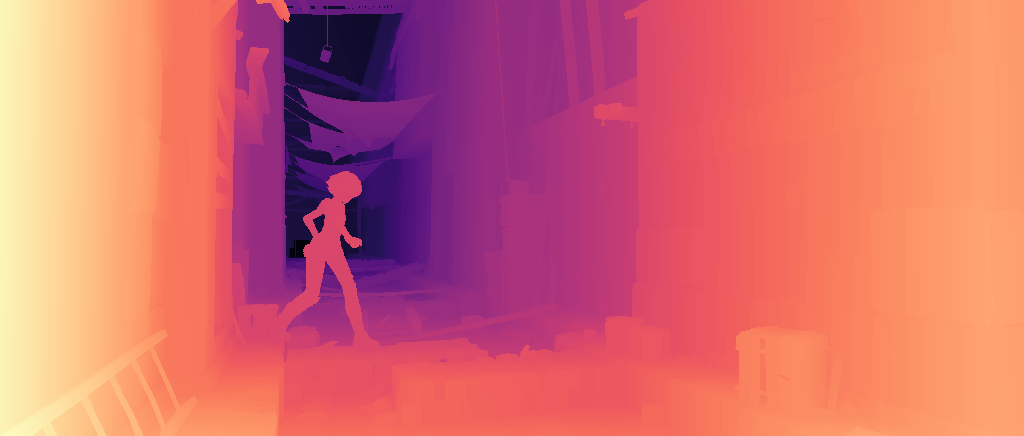}{0.35}{0.7}{0.15}{GT Depth} \\

    \showZoomCrop{imgs/sintel_highres_ablation_v2/3_rgb.png}{0.35}{0.7}{0.15}{} &
    \showZoomCrop{imgs/sintel_highres_ablation_v2/3_cfg1.png}{0.35}{0.7}{0.15}{} &
    \showZoomCrop{imgs/sintel_highres_ablation_v2/3_cfg2.png}{0.35}{0.7}{0.15}{} &
    \showZoomCrop{imgs/sintel_highres_ablation_v2/3_cfg3.png}{0.35}{0.7}{0.15}{} &
    \showZoomCrop{imgs/sintel_highres_ablation_v2/3_cfg4.png}{0.35}{0.7}{0.15}{} &
    \showZoomCrop{imgs/sintel_highres_ablation_v2/3_gt.png}{0.35}{0.7}{0.15}{} \\[0.5em]
    
    \drawZoomBox{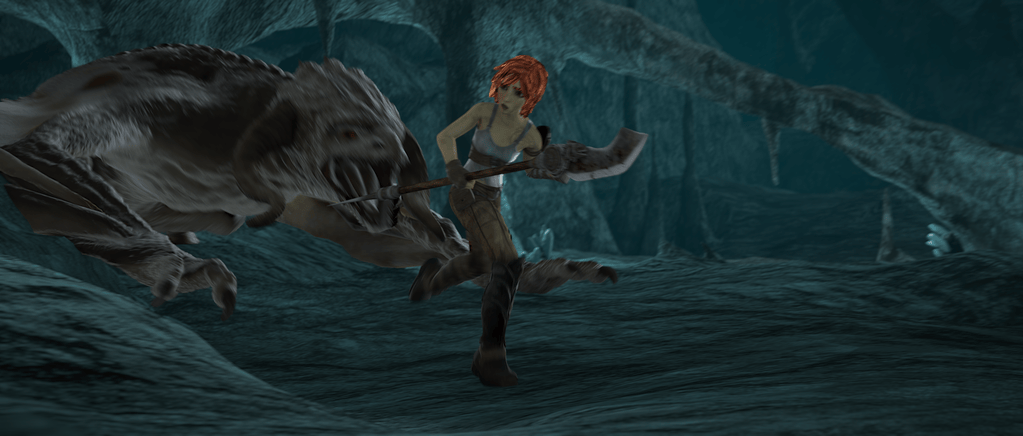}{0.65}{0.4}{0.15}{} &
    \drawZoomBox{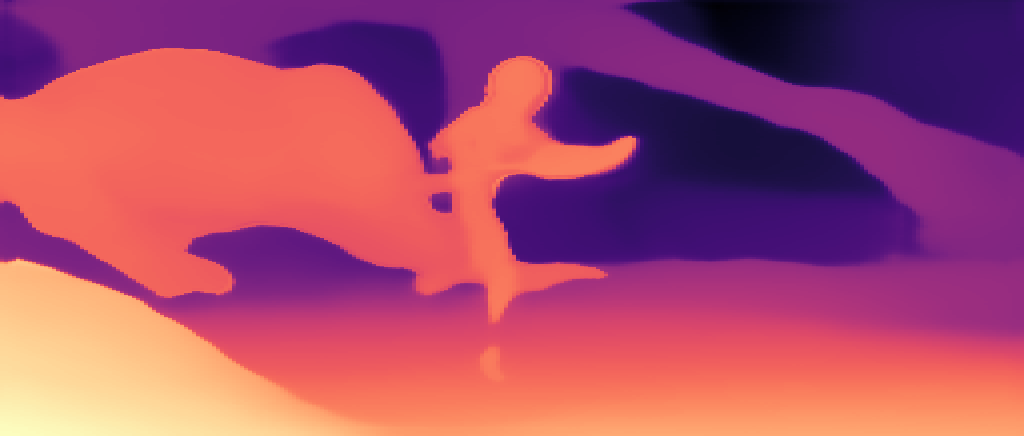}{0.65}{0.4}{0.15}{} &
    \drawZoomBox{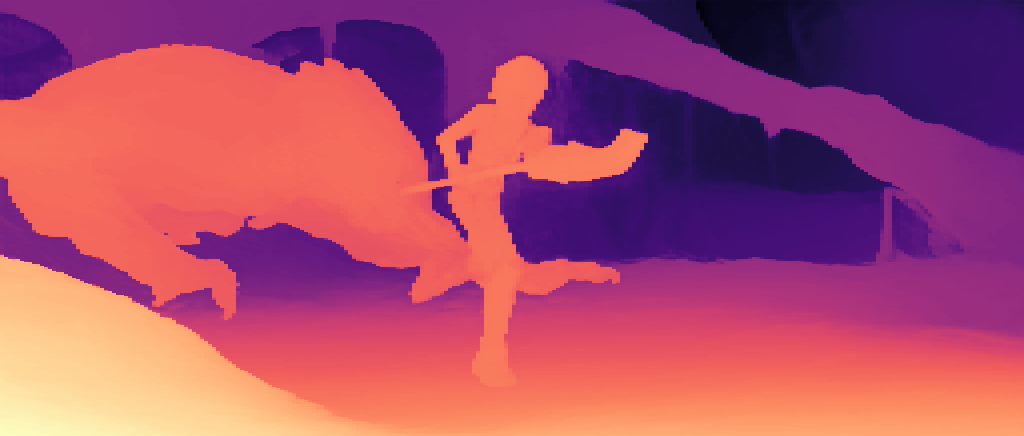}{0.65}{0.4}{0.15}{} &
    \drawZoomBox{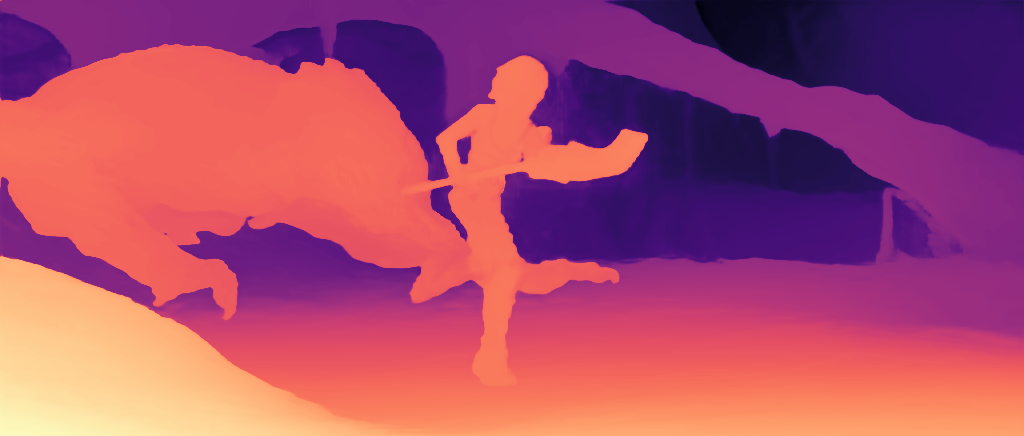}{0.65}{0.4}{0.15}{} &
    \drawZoomBox{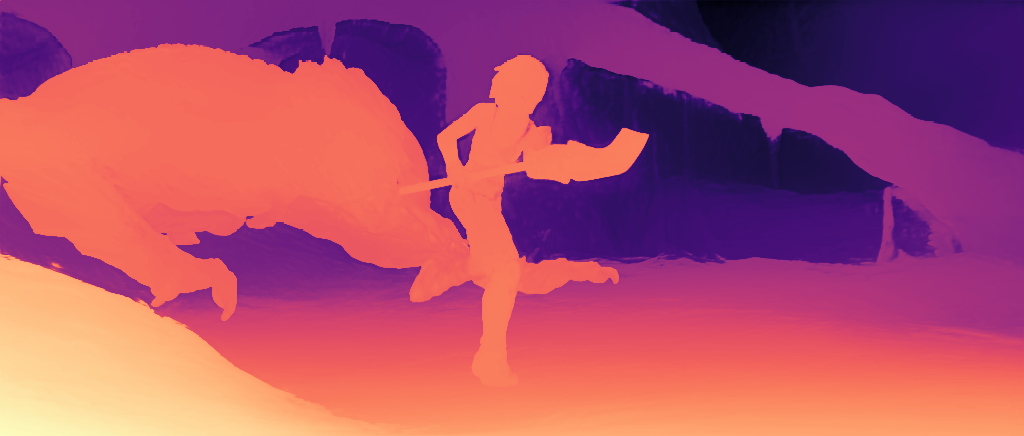}{0.65}{0.4}{0.15}{} &
    \drawZoomBox{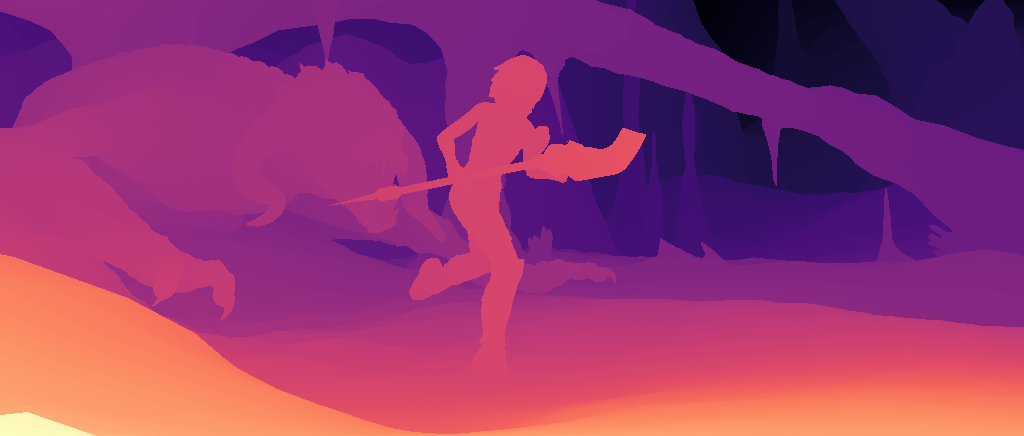}{0.65}{0.4}{0.15}{} \\

    \showZoomCrop{imgs/sintel_highres_ablation_v2/2_rgb.png}{0.65}{0.4}{0.15}{} &
    \showZoomCrop{imgs/sintel_highres_ablation_v2/2_cfg1.png}{0.65}{0.4}{0.15}{} &
    \showZoomCrop{imgs/sintel_highres_ablation_v2/2_cfg2.png}{0.65}{0.4}{0.15}{} &
    \showZoomCrop{imgs/sintel_highres_ablation_v2/2_cfg3.png}{0.65}{0.4}{0.15}{} &
    \showZoomCrop{imgs/sintel_highres_ablation_v2/2_cfg4.png}{0.65}{0.4}{0.15}{} &
    \showZoomCrop{imgs/sintel_highres_ablation_v2/2_gt.png}{0.65}{0.4}{0.15}{} \\[0.5em]
    
    \drawZoomBox{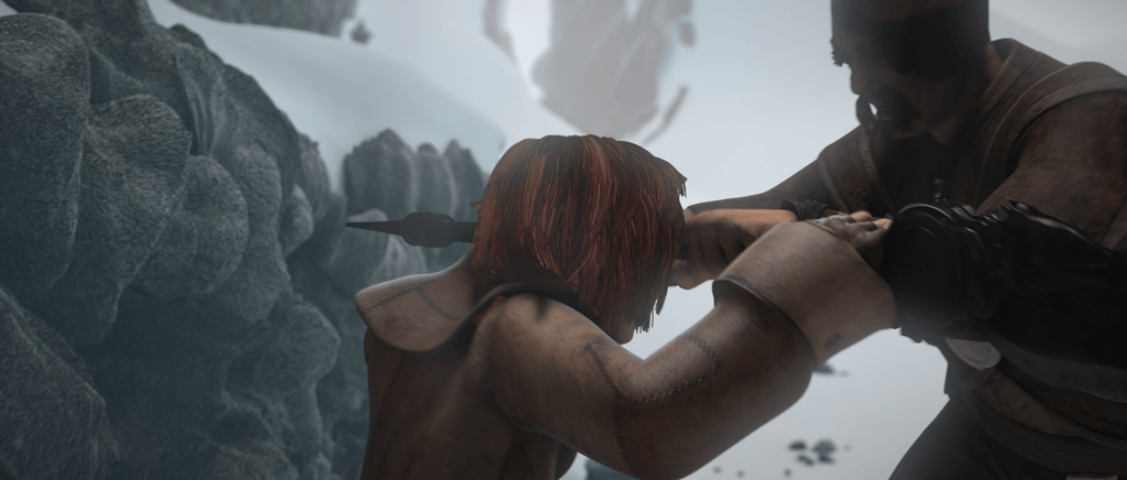}{0.35}{0.4}{0.15}{} &
    \drawZoomBox{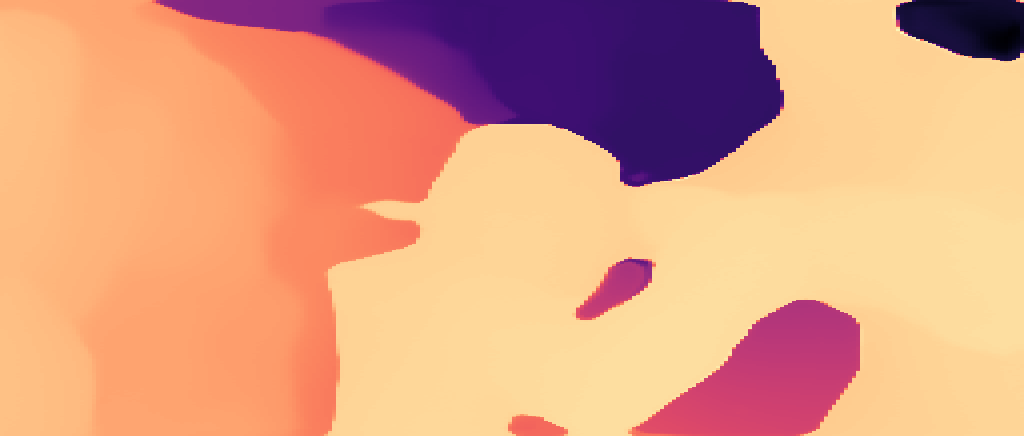}{0.35}{0.4}{0.15}{} &
    \drawZoomBox{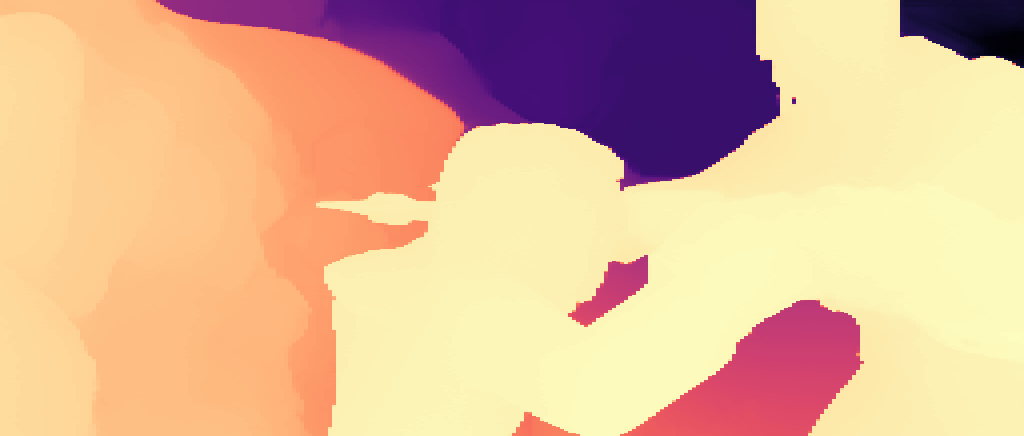}{0.35}{0.4}{0.15}{} &
    \drawZoomBox{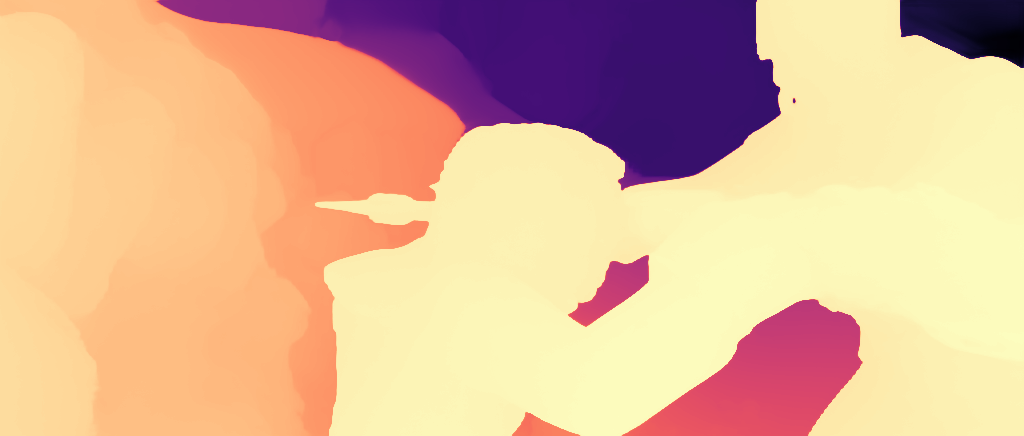}{0.35}{0.4}{0.15}{} &
    \drawZoomBox{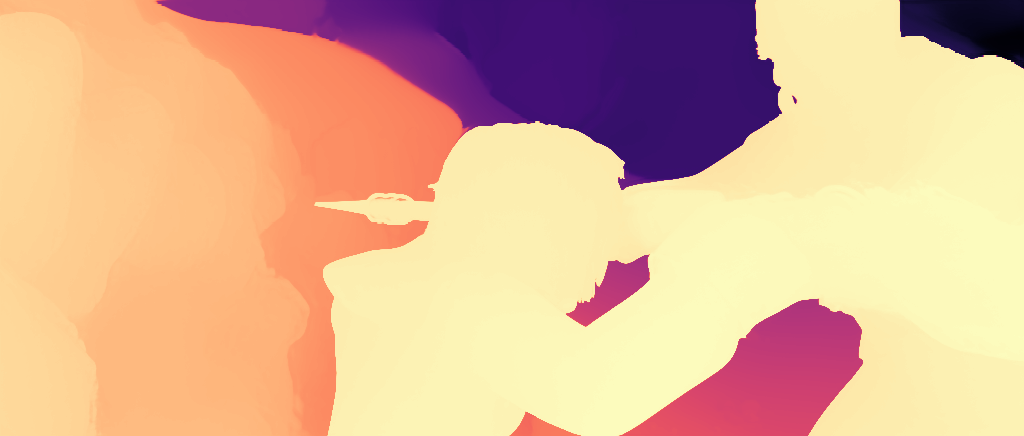}{0.35}{0.4}{0.15}{} &
    \drawZoomBox{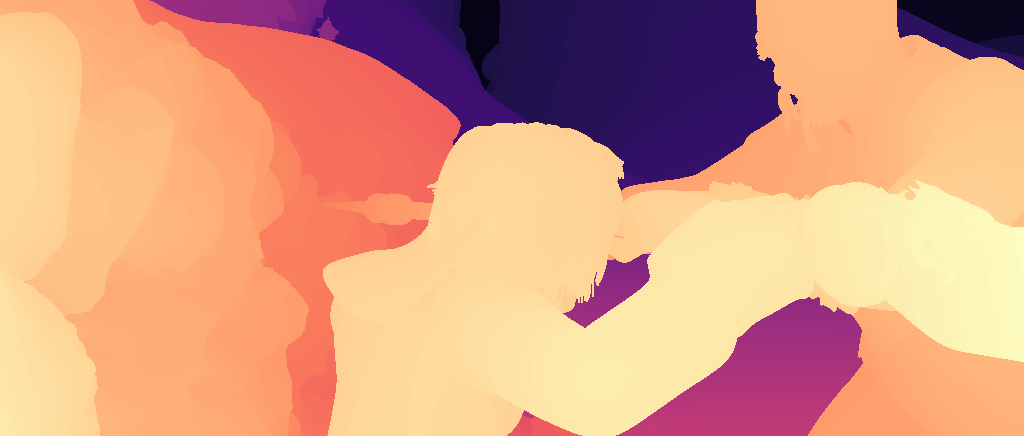}{0.35}{0.4}{0.15}{} \\

    \showZoomCrop{imgs/sintel_highres_ablation_v2/8_rgb.png}{0.35}{0.4}{0.15}{} &
    \showZoomCrop{imgs/sintel_highres_ablation_v2/8_cfg1.png}{0.35}{0.4}{0.15}{} &
    \showZoomCrop{imgs/sintel_highres_ablation_v2/8_cfg2.png}{0.35}{0.4}{0.15}{} &
    \showZoomCrop{imgs/sintel_highres_ablation_v2/8_cfg3.png}{0.35}{0.4}{0.15}{} &
    \showZoomCrop{imgs/sintel_highres_ablation_v2/8_cfg4.png}{0.35}{0.4}{0.15}{} &
    \showZoomCrop{imgs/sintel_highres_ablation_v2/8_gt.png}{0.35}{0.4}{0.15}{} \\
    
\end{tabular}
\caption{\textbf{Visualizing sub-pixel detail recovery} --
We propose a visual comparison of the different high-res configurations.
Config \raisebox{0.75pt}{\textcircled{\scriptsize 4}} achieves the highest fidelity, it preserves sharp edges and recovers fine details—such as the hair in the bottom row—without increasing the computational cost or memory requirements of the overall model.
}

\label{supfig:subpixel_dec}
\end{figure*}
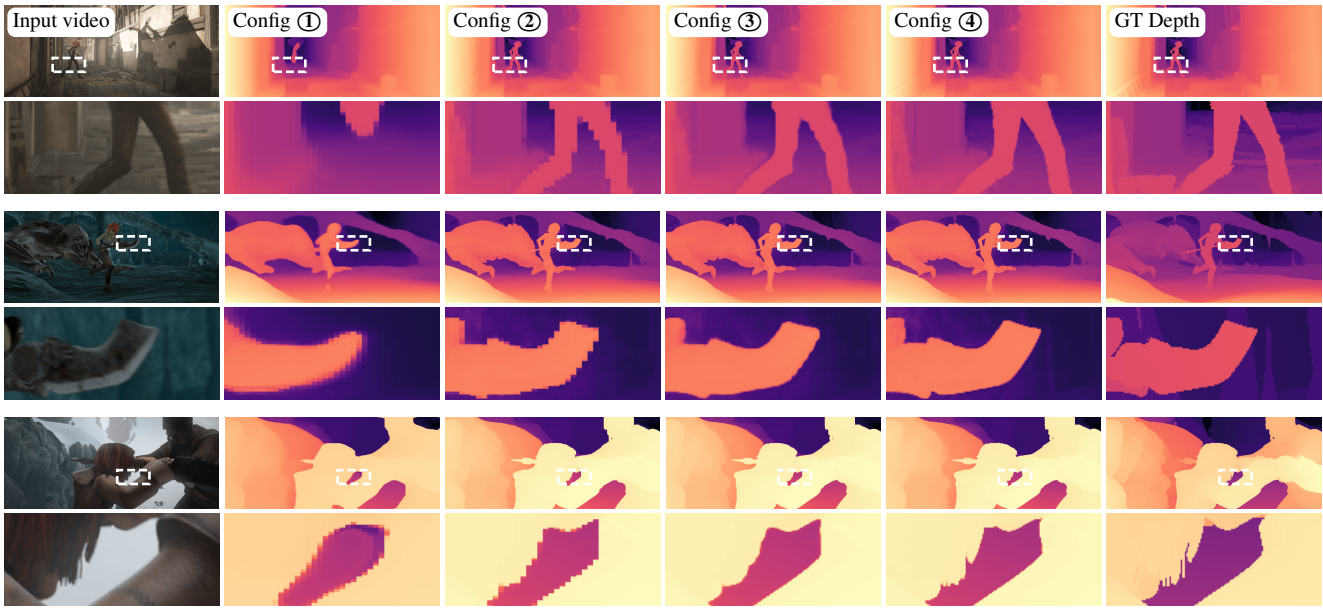

\section{Further Ablations}

\paragraph{Pretrained Encoder.}
In \cref{tab:app:abla_init}, we show our model performance when the video encoder is initialized with random weights compared to using pre-trained VideoMAE~\cite{videomaev2} weights.
We observe significant improvements across the board.

\begin{table}[h]
\centering
\small
\tabcolsep=0.4em
\begin{tabular}{@{}c cc ccc@{}}
\toprule
\multirow{2}[2]{*}{\textbf{\shortstack{Model weight initialization}}} & \multicolumn{2}{c}{\textbf{Video Depth Estimation}} & \multicolumn{3}{c}{\textbf{Camera Pose Estimation}} \\ 
\cmidrule(lr){2-3}\cmidrule(lr){4-6}
  & AbsRel (S)~$\downarrow$  & AbsRel (SS)~$\downarrow$ & ATE~$\downarrow$ & RPE-T~$\downarrow$& RPE-R~$\downarrow$ \\ 
\midrule
None  &  0.738 & 0.520 & 0.334 &   0.139 & 1.126 \\
VideoMAE~\cite{videomaev2}  & \best{0.302} & \best{0.257} & \best{0.091} & \best{0.028} & \best{0.245}  \\

\bottomrule
\end{tabular}

\caption{\textbf{Model initialization} --
Initializing the model with VideoMAE~\cite{videomaev2} weights leads to significant improvements.}
\label{tab:app:abla_init}
\end{table}

\paragraph{Local RGB patch size.}
We perform an ablation study on the size of the local RGB patch.
As shown in \cref{fig:ablation_patch_size}, our results indicate that the patch size 9$\times$9 yields the best overall performance across camera pose and depth estimation tasks.

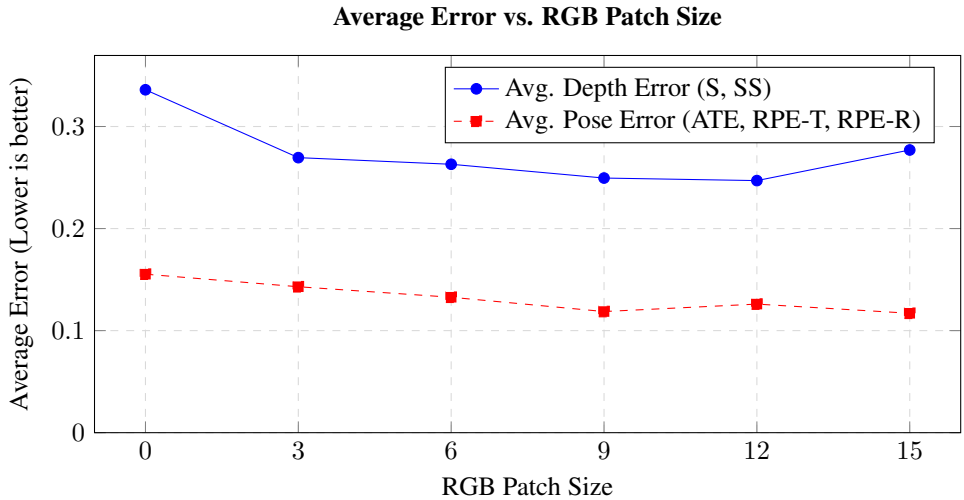
\begin{figure}[h]
\centering
\begin{tikzpicture}
\begin{axis}[
    title={\textbf{Average Error vs. RGB Patch Size}},
    width=0.75\linewidth, %
    height=6.6cm,
    xlabel={RGB Patch Size},
    ylabel={Average Error (Lower is better)},
    ylabel style={rotate=0}, %
    xtick={0, 3, 6, 9, 12, 15}, %
    grid=major, %
    grid style={dashed, gray!30}, %
    legend pos=north east, %
    legend cell align={left},
    ymin=0, %
    xmin=-1, %
    xmax=16,
]

\addplot[
    color=blue,       %
    mark=*,           %
    solid             %
] coordinates {
    (0, 0.3360)
    (3, 0.2695)
    (6, 0.2630)
    (9, 0.2495)
    (12, 0.2470)
    (15, 0.2770)
};
\addlegendentry{Avg. Depth Error (S, SS)}

\addplot[
    color=red,        %
    mark=square*,     %
    dashed            %
] coordinates {
    (0, 0.1553)
    (3, 0.1430)
    (6, 0.1327)
    (9, 0.1187)
    (12, 0.1260)
    (15, 0.1170)
};
\addlegendentry{Avg. Pose Error (ATE, RPE-T, RPE-R)}

\end{axis}
\end{tikzpicture}
\caption{\textbf{Ablation study on RGB patch size} -- The plot shows the average error for depth and pose estimation on Sintel. A patch size between 9 and 12 yields the best performance for both tasks.}
\label{fig:ablation_patch_size}
\end{figure}

\section{Further Qualitative Results}
Complementing the qualitative visualizations in the main text, we provide additional examples of our reconstruction results in \cref{supfig:ours_vis}, along with further baseline comparisons in \cref{supfig:side_by_side_vis} and \cref{supfig:depth_vis}.

\begin{figure*}[h]
\centering
\vspace{2mm}
\includegraphics[width=\linewidth]{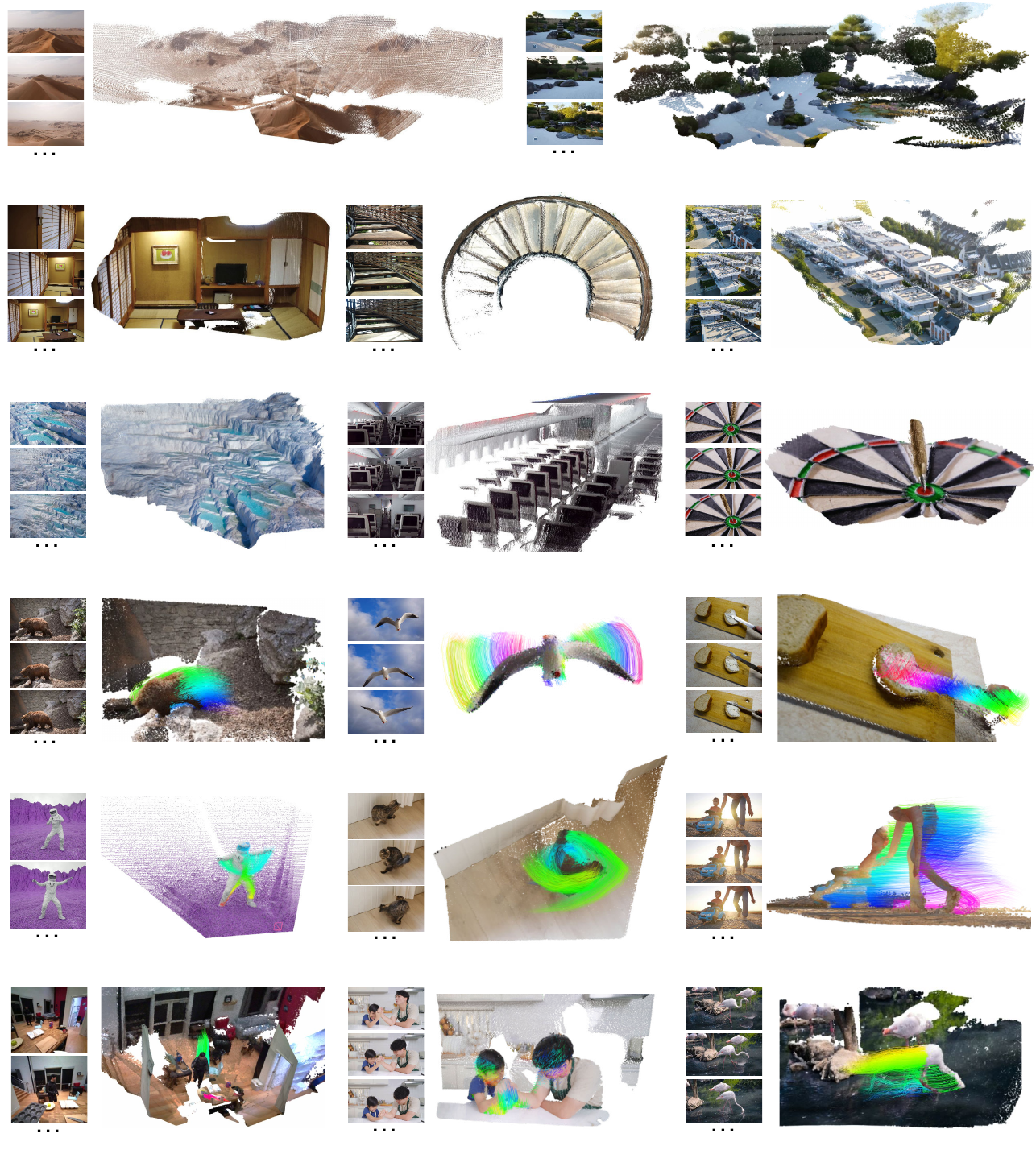}
\vspace{-3mm}
\caption{
\textbf{Additional visualizations of \name} --  \name produces accurate reconstructions for both static environments (top three rows) and dynamic sequences (bottom three rows).}
\label{supfig:ours_vis}
\end{figure*}

\begin{figure*}[h]
\centering
\small
\tabcolsep=0em
\noindent\begin{tabular}{
  >{\centering\arraybackslash}p{0.09\linewidth}
  >{\centering\arraybackslash}p{0.2275\linewidth}
  >{\centering\arraybackslash}p{0.2275\linewidth}
  >{\centering\arraybackslash}p{0.2275\linewidth}
  >{\centering\arraybackslash}p{0.2275\linewidth}
}
{Input video} & ~~~~~~MegaSaM~\cite{li2025megasam} & ~~~$\pi^3$~\cite{wang2025pi} & ~~SpatialTrackerV2~\cite{xiao2025spatialtrackerv2} & \textbf{D4RT (Ours)} \\[5pt]
\end{tabular}
\includegraphics[width=1.0\linewidth, trim=0 0 0 1cm, clip]{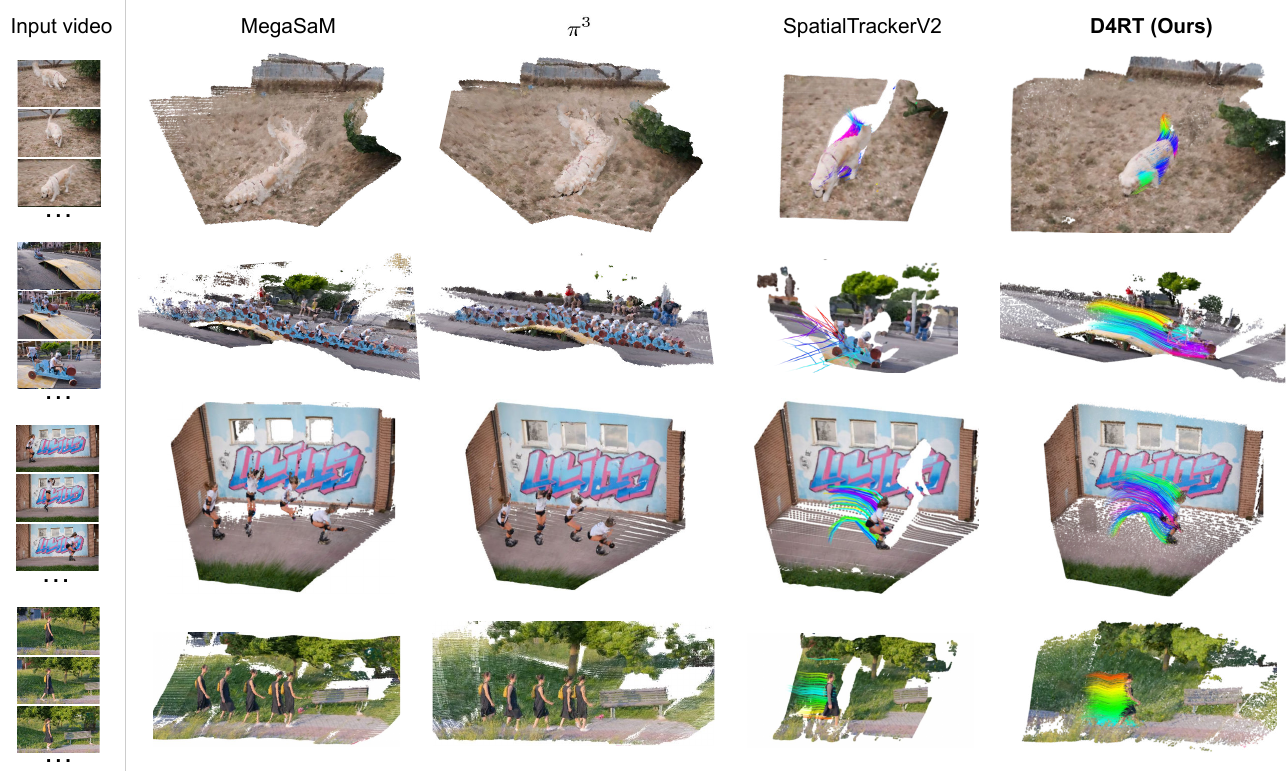}
\vspace{-5mm}
\caption{
\textbf{Additional reconstruction results across methods} -- Pure reconstruction methods (MegaSaM and $\pi^3$) are visualized as accumulated point clouds. For SpatialTrackerV2, we visualize sparse tracks on a representative frame. In contrast, \name reconstructs a complete 4D scene representation, tracking \emph{all} pixels across the entire video.
}
\vspace{3mm}
\label{supfig:side_by_side_vis}
\end{figure*}

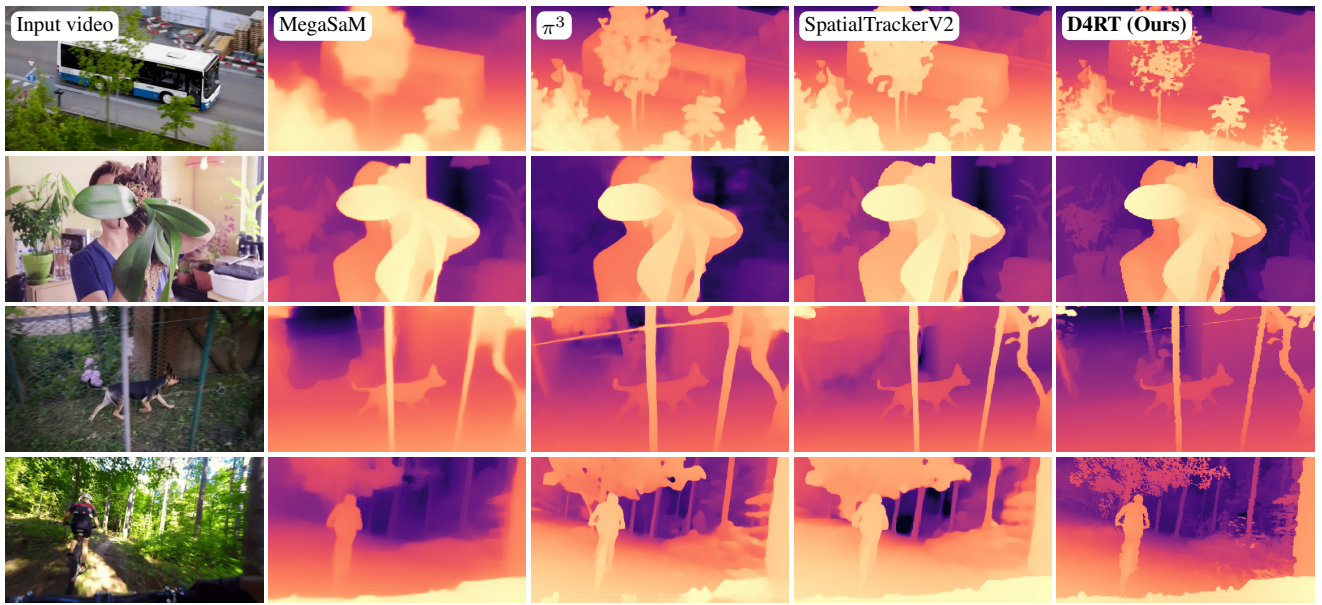
\begin{figure*}[h]
\centering
\usetikzlibrary{calc} 

\newcommand{\labeledImage}[2]{%
    \begin{tikzpicture}
        \node[anchor=south west, inner sep=0] (image) at (0,0) {
            \includegraphics[width=0.196\linewidth]{#2}
        };
        \node[anchor=north west, fill=white, rounded corners=3pt, inner sep=2pt, font=\footnotesize, text=black] 
            at ([xshift=2pt, yshift=-2pt]image.north west) {#1};
    \end{tikzpicture}%
}

\setlength{\tabcolsep}{1pt}
\renewcommand{\arraystretch}{0.6} %

\begin{tabular}{@{}ccccc@{}}
    
    \labeledImage{Input video}{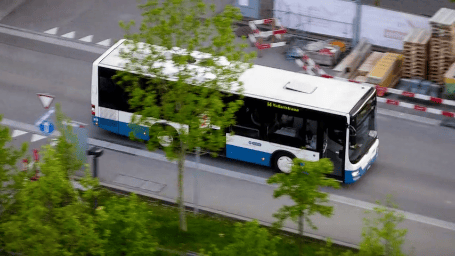} &
    \labeledImage{MegaSaM}{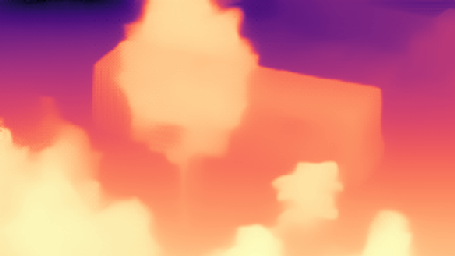} &
    \labeledImage{$\pi^3$}{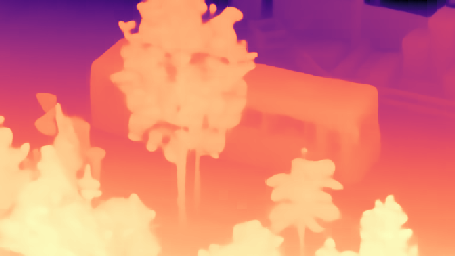} &
    \labeledImage{SpatialTrackerV2}{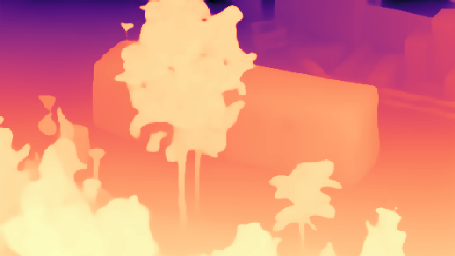} &
    \labeledImage{\textbf{D4RT (Ours)}}{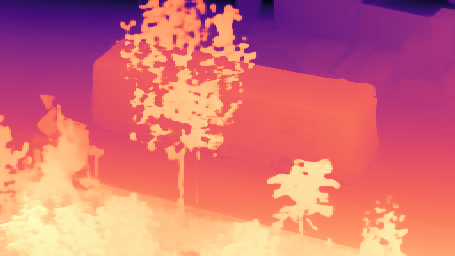} \\
    \includegraphics[width=0.196\linewidth]{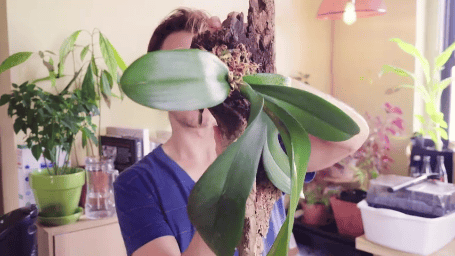} &
    \includegraphics[width=0.196\linewidth]{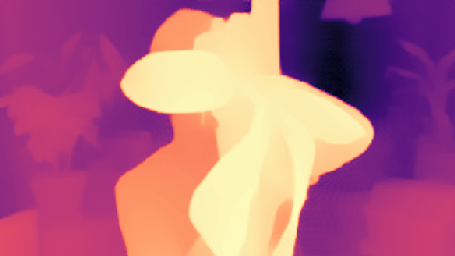} &
    \includegraphics[width=0.196\linewidth]{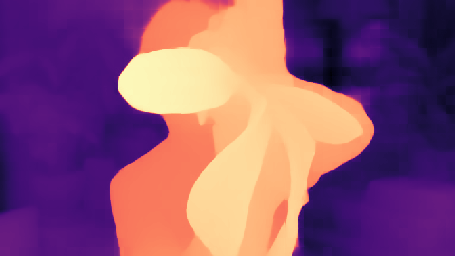} &
    \includegraphics[width=0.196\linewidth]{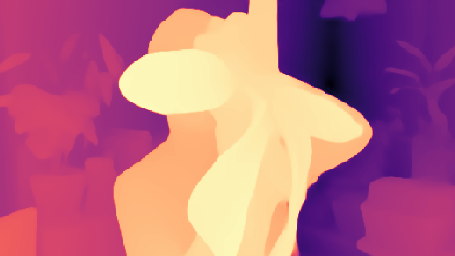} &
    \includegraphics[width=0.196\linewidth]{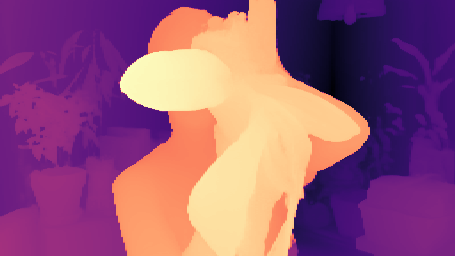} \\
    \includegraphics[width=0.196\linewidth]{imgs/davis17_depth_comparison/libby_rgb_16.png} &
    \includegraphics[width=0.196\linewidth]{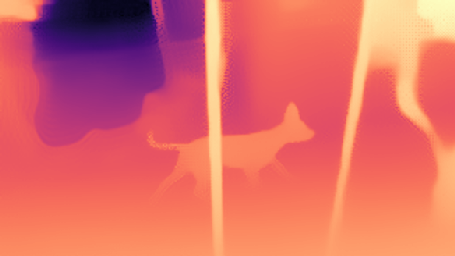} &
    \includegraphics[width=0.196\linewidth]{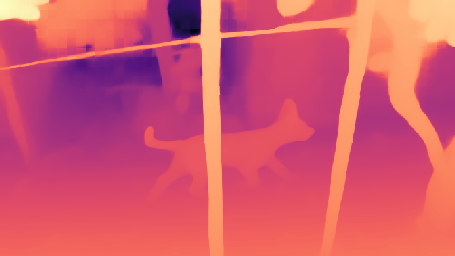} &
    \includegraphics[width=0.196\linewidth]{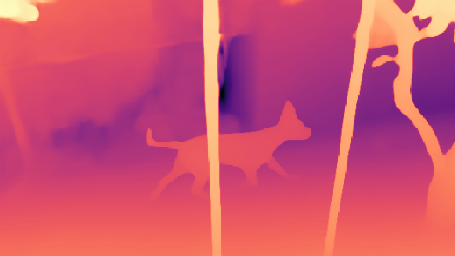} &
    \includegraphics[width=0.196\linewidth]{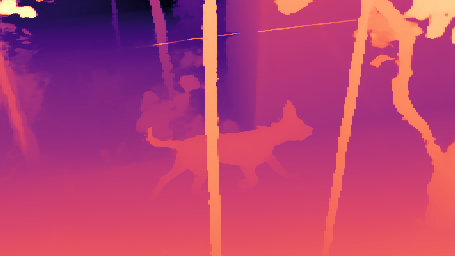} \\
    \includegraphics[width=0.196\linewidth]{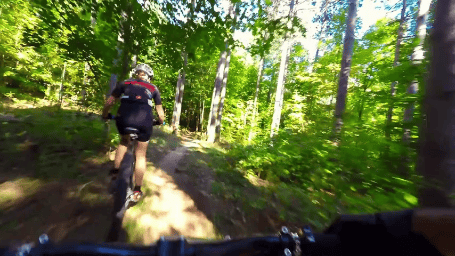} &
    \includegraphics[width=0.196\linewidth]{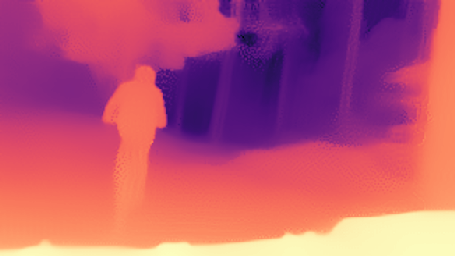} &
    \includegraphics[width=0.196\linewidth]{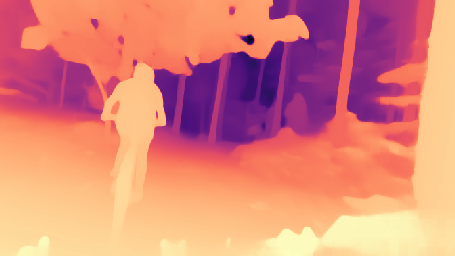} &
    \includegraphics[width=0.196\linewidth]{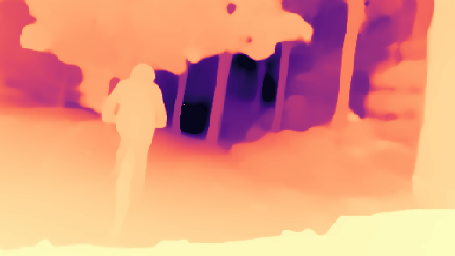} &
    \includegraphics[width=0.196\linewidth]{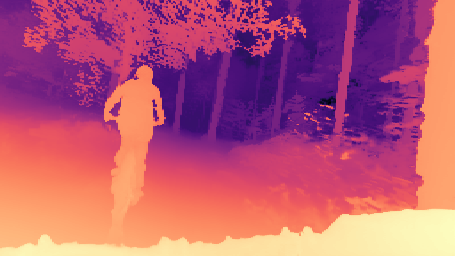} \\
    
\end{tabular}
\caption{
\textbf{Qualitative depth comparison across methods} --
    \name is able to perform dense depth estimation with finer details than current state-of-the-art methods, preserving geometric accuracy even in scenarios with large motion blur.
}
\label{supfig:depth_vis}
\end{figure*}

\end{document}